%% file: main_a.tex
\documentclass[final,1p,times]{elsarticle}

\usepackage{graphics} 
\usepackage{epsfig} 
\usepackage{times} 
\usepackage{amsmath} 
\usepackage{amssymb}  
\usepackage[table]{xcolor}
\usepackage[ruled,linesnumbered, noend]{algorithm2e}

\usepackage{adjustbox}
\usepackage{comment}
\usepackage{multirow}
\usepackage{graphicx}
\usepackage{wrapfig}
\usepackage{subfigure}
\usepackage{balance}
\usepackage{wrapfig}
\usepackage{booktabs}  
\usepackage{float}

\usepackage{soul, color}
\usepackage{colortbl}
\definecolor{Gray}{gray}{0.9}
\definecolor{Yellow}{RGB}{255 255 0}

\usepackage{pifont}
\newcommand{\xmark}{\ding{55}}%

\newcommand{\ve}[1]{\mathbf{#1}} 
\newcommand{\hve}[1]{\hat{\mathbf{#1}}} 
\newcommand{\tve}[1]{\tilde{\mathbf{#1}}} 

\journal{}

\begin{document}

\begin{frontmatter}



\title{Real-time Human Finger Pointing Recognition and Estimation for Robot Directives Using a Single Web-Camera}


\author[label1]{Eran Beeri Bamani} 
\author[label2]{Eden Nissinman} 
\author[label2]{Lisa Koenigsberg} 
\author[label2]{Inbar Meir} 
\author[label2]{Yoav Matalon} 
\author[label2]{Avishai Sintov\corref{cor1}}
\affiliation[label1]{organization={Department of Mechanical Engineering, Massachusetts Institute of Technology (MIT)},
            addressline={},
            city={Cambridge},
            postcode={02139},
            country={MA, USA}}
\affiliation[label2]{organization={School of Mechanical Engineering, Tel-Aviv University},
            addressline={Haim Levanon St.},
            city={Tel-Aviv},
            postcode={6997801},
            country={Israel}}
\ead{sintov1@tauex.tau.ac.il}
\cortext[cor1]{Corresponding Author.}
%





\input{abstract}


\end{frontmatter}

\section{Introduction}
\input{Introduction}

\label{sec:introduction}

\section{Related Work}
\input{relatedwork}

\label{sec:relatedwork}

\section{Preliminary Pointing Analysis}
\input{pre}
\label{sec:preliminary}

\section{Methods}
\input{Method}
\label{sec:method}

\section{Model Evaluation}
\input{Evaluation}

\label{sec:Evaluation}

\section{Robot Experiments}
\input{Experiments}
\label{sec:experiments}

\section{Conclusions}
\input{Conclusions}

\section*{Funding}

This work was supported by the Israel Innovation Authority (grant No. 77857).


\bibliographystyle{elsarticle-num} 
\bibliography{ref}

\end{document}

%% file: Abstract.tex
\begin{abstract}
    Gestures play a pivotal role in human communication, often serving as a preferred or complementary medium to verbal expression due to their superior spatial reference capabilities. A finger-pointing gesture conveys vital information regarding some point of interest in the environment. In Human-Robot Interaction (HRI), users can easily direct robots to target locations, facilitating tasks in diverse domains such as search and rescue or factory assistance. State-of-the-art approaches for visual pointing estimation often rely on depth cameras, are limited to indoor environments, and provide discrete predictions between limited targets. In this paper, we explore the development of models that enable robots to understand pointing directives from humans using a single web camera, even in diverse indoor and outdoor environments. A novel perception framework is proposed which includes a designated data-based model termed \textit{PointingNet}. PointingNet recognizes the occurrence of pointing through classification followed by approximating the position and direction of the index finger with an advanced regression model. The model relies on a novel segmentation model for masking any lifted arm. While state-of-the-art human pose estimation models provide poor pointing angle estimation error of 28$^\circ$, PointingNet exhibits a mean error of less than 2$^\circ$. With the pointing information, the target location is computed, followed by robot motion planning and execution. The framework is evaluated on two robotic systems, demonstrating accurate target reaching.

\end{abstract}

%% file: Introduction.tex

Finger pointing is a universal gesture in which a human extends the arm and index finger toward a desired target \cite{Cooperrider2018}. This gesture serves as a prominent form of communication across various cultures, enabling individuals to direct attention to specific regions of interest in their environment \cite{Kita2003}. Tomasello \cite{Tomasello2008} suggested that gestures, particularly pointing, preceded vocal communication due to their superior ability to convey spatial references. Unlike speech, pointing naturally and efficiently directs human attention to external targets. Infants, for instance, use pointing before they learn to pronounce words \cite{Bates2002}. Furthermore, in multi-cultural interaction, pointing is the preferred form of communication to cope with the language barrier \cite{Hewes1974}. Consequently, pointing is an indispensable tool for conveying important information. Pointing recognition in Human-Robot Interaction (HRI), therefore, is a necessary form of communication for natural and easy directives to robots.
\begin{figure}
    \centering
    \includegraphics[width=\linewidth]{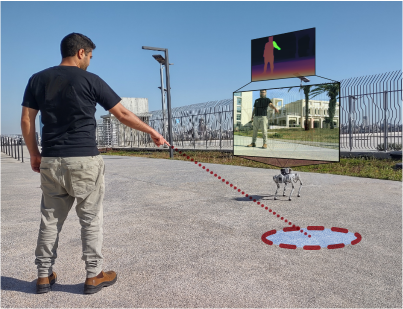}
    \caption{A user directs a quadruped robot to a target position by pointing. The robot observes the user through a web camera. PointingNet  identifies a pointing gesture and estimates its position and direction. Once the target has been calculated, the robot plans motion and moves to the target.}
    \label{fig:withGo1}
    \vspace{-0.5cm}
\end{figure}

HRI is a broad field of study focused on the exploration of mutual interactions between humans and robots. Such interactions can take various forms, including information exchange, collaboration, or guidance. Gesture recognition, as a non-verbal communication approach, has been a popular method in HRI and has traditionally relied on conventional classification techniques utilizing handcrafted features \cite{lee1999hmm, dardas2011real, hussein2013human}. However, these methods often struggle to address the complexity and variability of human motions. As a result, considerable effort has been directed toward modeling and understanding human intent through natural gestures \cite{Ji2020,Haque2023}. By leveraging natural gestures, particularly pointing, humans can effortlessly issue directive instructions to robots, thereby reducing the cognitive load on users \cite{Villani2018}. This reduction in cognitive effort eliminates the need for users to memorize complex commands or navigate through intricate digital interfaces. For example, first responders can efficiently direct a robot to assist in life-saving tasks, while medical personnel can instruct a robot to retrieve tools or supplies seamlessly.

\begin{figure*}
    \centering
    \includegraphics[width=\linewidth]{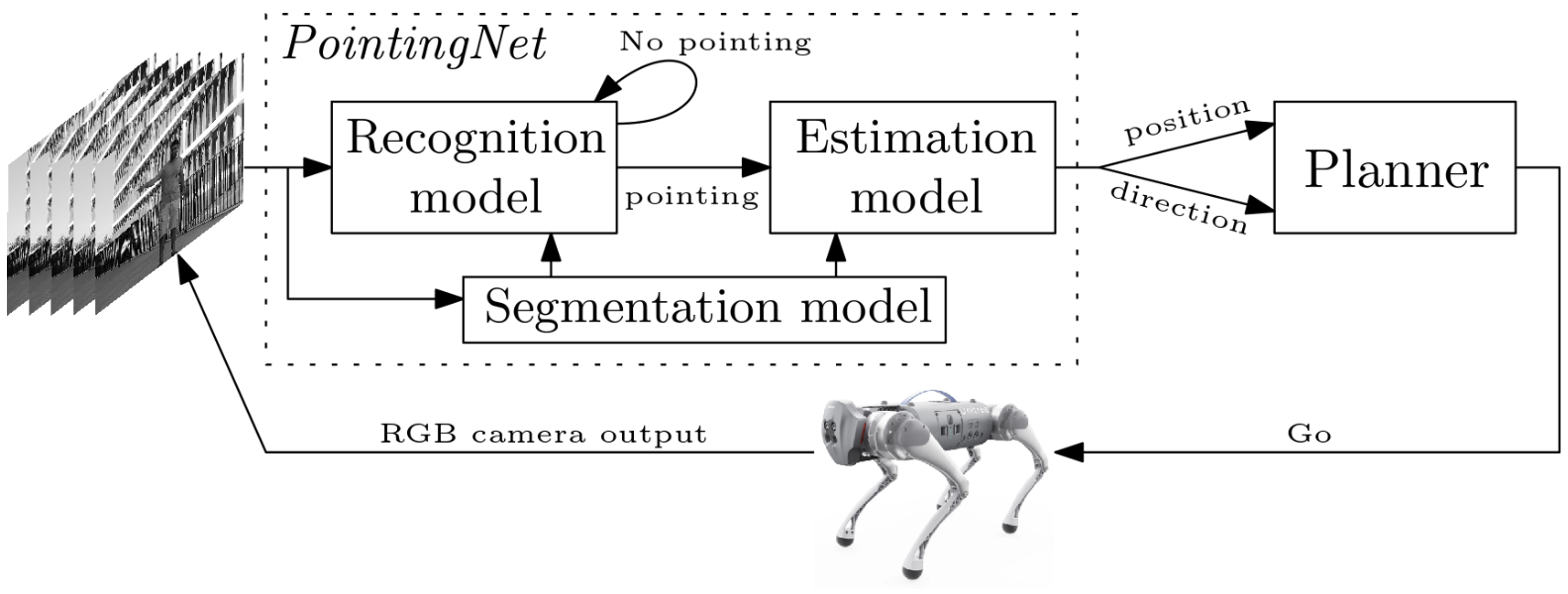}
    \caption{Illustration of the proposed framework for a robot to reach a pointed target. Pointing recognition acts as a trigger for prompting directive motion. As long as the robot does not recognize a pointing gesture, it will remain in an idling mode. Once pointing of the user has been identified, the pointing position and direction are estimated. The estimations are given with respect to the coordinate frame of the camera on the robot. Hence, they are used by a motion planner for planning the motion of the robot from its current pose to the directed target.}
    \label{fig:robot_scheme}
\end{figure*}
For a robot to interpret a human's pointing gesture, it must first recognize the occurrence of the gesture and subsequently estimate its direction. While various approaches have been proposed for estimating pointing direction, fewer studies have addressed the recognition aspect. Direction estimation methods can generally be categorized based on the sensing modality into two main approaches: wearable devices and visual perception. Wearable devices \cite{Haque2015, Sikeridis2017}, further discussed in the next section, are often cumbersome and prevent spontaneous pointing gestures by occasional users.
In contrast, visual perception enables spontaneous detection of pointing gestures by observing frequent or occasional users through one or more cameras. Significant progress has been made in pointing detection using multiple RGB cameras or depth (stereo) cameras \cite{Littmann1996, Park2011,nagai2017finger,Surendran2023,Serpiva2024}. Although depth cameras facilitate easier detection of pointing gestures, their application is constrained to short-range, indoor environments, and their performance deteriorates in outdoor settings. Additionally, they may require specialized hardware setups, which can limit the versatility and generalizability of the method \cite{Jirak2020}.

Only a limited number of studies have addressed the pointing gesture problem using a single RGB camera \cite{Jirak2020}. These approaches are typically demonstrated in constrained settings, providing coarse estimations of pointing direction towards distinct regions with large, well-defined objects. Human pose estimation models, usually termed \textit{Skeleton} models \cite{Osokin2018}, are often employed to identify key-points that represent the pointing direction \cite{Medeiros2021}. However, these methods predominantly focus on the human forearm, with the direction being extracted analytically through calibration. As evaluated in this study, such models yield poor results and fail to achieve the level of accuracy required for precise pointing direction prediction.


To accurately infer a user's intended target, a robot must precisely estimate the position and direction of the user's finger relative to its own coordinate frame. This is critical in environments with low diversity, where even minor errors can lead to mission failure, such as directing a search and rescue drone to a specific window \cite{Deuerlein2020}. While various approaches use the forearm \cite{Haque2015}, index finger \cite{Hu2010}, or a nose-to-finger vector \cite{Azari2019} to measure pointing, no comprehensive analysis has determined which method best captures a user's intent. Most existing work relies on forearm direction \cite{Lai2016,Kuramochi2021,Constantin2022}, yet the forearm and index finger are not always co-linear. Previous efforts to estimate index finger direction are limited to discrete scenarios, often solving classification problems to distinguish between a few objects on an indoor tabletop \cite{Hu2010, Shukla2015,Constantin2022}. In summary, current methods have significant limitations, including dependence on restrictive skeleton models, being confined to indoor spaces, using discrete spatial representations, and requiring RGB-D cameras. 
To the best of the authors' knowledge, no prior work has addressed the challenge of continuous, real-time pointing estimation using a single RGB camera across both indoor and outdoor settings, while simultaneously incorporating pointing detection.

In this paper, we aim to enable a mobile robot to recognize pointing gestures, estimate the target location, and navigate toward it (Figure \ref{fig:withGo1}). We explore the development of robust models for pointing recognition and estimation in diverse environments. First, we present a comprehensive experimental analysis to evaluate the accuracy of three distinct pointing measurement approaches. Subsequently, we propose a complete framework for accurate pointing recognition and estimation based solely on RGB images. The framework, illustrated in Figure \ref{fig:robot_scheme}, introduces a novel model termed PointingNet, which integrates two primary components: a pointing recognition model and a pointing estimation model. Both models are based on a unique segmentation model aimed to mask any visibly lifted arm within the scene. The recognition model classifies the segmented arm to determine whether a pointing gesture is present. If pointing is detected, a novel estimation model is activated to approximate the position and direction of the index finger relative to the camera. Based on this information, the robot computes the desired target location and moves toward it.

To summarize, the key contributions of this work are as follows:
\begin{itemize}
    \item This work introduces a novel method for continuous, real-time pointing estimation that, unlike prior work, is designed for and evaluated in unstructured indoor and outdoor environments using only a single RGB camera. 
    \item The framework integrates a robust recognition model that achieves high success rates in identifying pointing gestures.
    \item A novel pointing estimation model is introduced to estimate the position and direction of the index finger during pointing gestures. Evaluation on a diverse test set demonstrates high accuracy in these estimations.
    \item We conduct a comparative analysis of different pointing direction measurement approaches, focusing on the intended target of the user.
    \item A unique segmentation model is employed to isolate and mask the pointing arm, enabling precise recognition and estimation. This segmentation model can also be applied to other gesture recognition applications.
    \item The trained model and a labeled pointing dataset are made available as an open-source resource for the research community\footnote{The dataset will be available upon request, with images modified to protect participant privacy.}.
\end{itemize}
The proposed method has significant potential across various industries, especially in automation and inspection. By allowing operators to guide robots with natural pointing gestures, it streamlines tasks in manufacturing and logistics. Its adaptability to diverse indoor and outdoor environments makes it ideal for warehouse management and construction. Beyond industrial use, the technology can enhance virtual and augmented reality, interactive presentations, and gaming, enabling more intuitive human-computer interaction. The open-source availability of the method and its dataset will further accelerate its adoption and advance human-robot collaboration across various domains. While designed for a robot-mounted camera, it can also be used with a stationary camera to direct a remote robot.

%% file: relatedwork.tex
\begin{table*}[]
    \centering
    \caption{State-of-the-art comparison of pointing recognition methods}
    \label{tb:sota}
    \begin{adjustbox}{width=\linewidth}
    \begin{tabular}{lc|cccccccc}\toprule
        \multirow{2}{*}{Paper} && Recognition & Body & Skeleton & IF/FA/EF & Indoor & Outdoor & RGB/Depth/ & \textbf{C}ontinuous/   \\     
         && & posture &&&&  & Wearable & \textbf{D}iscrete   \\\midrule
        
        Sikeridis et al. &
        \cite{Sikeridis2017} & \xmark & \xmark & - & FA & - & - & Wearable & C \\
        
        Haque et al. &
        \cite{Haque2015} & \checkmark & \xmark & - & FA & - & - & Wearable & C \\

        Surendran and Wagner & \cite{Surendran2023} & \xmark & \checkmark & \checkmark & FA/EF & \checkmark & \xmark & Depth & D \\
        
        Jirak et al. &
        \cite{Jirak2020} & \checkmark & \xmark & \xmark & IF & \checkmark & \xmark & RGB  & D  \\
        
        Medeiros et al. &
        \cite{Medeiros2021} & \xmark & \checkmark & \checkmark & FA & \xmark & \checkmark & RGB & D \\ 

        Hu et a. &
        \cite{Hu2010} & \xmark & \xmark & \xmark & IF & \checkmark & \xmark & 2$\times$RGB & C  \\

        Azari et al. &
        \cite{Azari2019} & \xmark & \checkmark & \xmark & EF & \checkmark & \xmark & Depth & C \\
        
        Lai et al. &
        \cite{Lai2016} & \checkmark & \checkmark & \checkmark & FA & \checkmark & \xmark & Depth & C \\
        
        Kuramochi et al. &
        \cite{Kuramochi2021} & \xmark & \xmark & \checkmark & FA & \checkmark & \xmark & 2$\times$RGB & C \\

        Constantin et al. &
        \cite{Constantin2022} & \checkmark & \xmark & \checkmark & FA & \checkmark & \xmark & RGB & D \\
        
        Shukla et al. &
        \cite{Shukla2015} & \xmark & \xmark & \xmark & IF & \checkmark & \xmark & Depth & D \\ 

        Nickel and Stiefelhagen &
        \cite{Nickel2007} & \checkmark & \checkmark & \xmark & FA/EF &  \checkmark & \xmark & Depth & C \\
        
        Martin et al. &
        \cite{Martin2010} & \xmark & \checkmark & \xmark & - & \checkmark & \xmark & RGB & D \\

        T{\"o}lgyessy et al. &
        \cite{Tlgyessy2017} & \xmark & \xmark & \checkmark & FA & \checkmark & \xmark & Depth & C \\ 
        
        Pietroszek et al. &
        \cite{Pietroszek2017} & \xmark & \xmark & - & FA & - & - & Wearable & C \\
        
        Das &
        \cite{Das2018} & \xmark & \xmark & \xmark & IF & \checkmark & \xmark & Depth & C \\
        

        Hu et al. &
        \cite{Hu2022} & \checkmark & \checkmark & \checkmark & FA & \checkmark & \xmark & Depth & D \\
        
        Kehl and Van Gool &
        \cite{Kehl2004} & \checkmark & \checkmark & \xmark & EF & \checkmark & \xmark & 8$\times$RGB & C \\

        Nakamura et al. & \cite{Nakamura2023} & \checkmark & \checkmark & \checkmark & FA & \checkmark & \xmark & 2$\times$RGB & C \\

        Pelgrim et al. & \cite{pelgrim2024} & \xmark & \checkmark & \checkmark & FA/EF & \checkmark & \xmark & Depth & D \\

        Lorentz et al. & \cite{Lorentz2023} & \xmark & \checkmark & \checkmark & - & \checkmark & \xmark & RGB & C \\

        \hline
        Proposed method && \checkmark & \xmark & \xmark & IF & \checkmark & \checkmark & RGB & C \\    
        \bottomrule
        
    \end{tabular}
    \end{adjustbox}
\end{table*}


In this section, we review the existing literature on pointing detection and estimation, relying to the three main groups of sensing technology: depth cameras, RGB cameras and wearable devices.

\subsection{Visual Perception with Depth Cameras}
The majority of research in visual perception has focused on depth cameras or multiple RGB cameras positioned to capture different views of the pointing gesture \cite{pelgrim2024}. Early studies utilized depth camera images to estimate pointing targets on a table \cite{Littmann1996}. Nickel and Stiefelhagen \cite{Nickel2007} used a prior probabilistic body model and clustering of the human skin color \cite{Jie1997} to observe the head and hands, and identify the direction of pointing gestures. Their approach also incorporated a Hidden Markov Model to detect the occurrence of pointing gestures. Hu et al. \cite{Hu2010} used two orthogonal camera views to detect and approximate pointing direction. More recently, Kuramochi et al. \cite{Kuramochi2021} employed two fish-eye cameras installed on either side of a display to capture images of a wide area in front of the display. This setup enabled skeleton recognition and estimation of pointing direction toward the display. Similarly, Zhou et al. \cite{Zhou2021} utilized contour detection and the Convex-Hull (CH) of the hand with close-up Kinect-based hand images. However, the applicability of such methods to real-world scenarios remains unclear. As discussed earlier, depth cameras are inherently limited to indoor environments, restricting the generalizability of these approaches.

\subsection{Visual perception with a single RGB camera}
Some studies have explored pointing detection using a single RGB camera. Martin et al. \cite{Martin2010} employed a simple monocular camera combined with a classifier to predict targets on an indoor floor from a set of discrete points within the user's proximal surroundings. Features from both the arm and head were extracted, and the estimation process was initiated through voice commands. Jirak et al. \cite{Jirak2020} focused on detecting whether the hand was pointing at objects, using a camera positioned above a table in a controlled environment. The Convex-Hull (CH) of the hand was utilized to verify that exactly one finger was extended. However, this method was constrained to detection from a specific camera angle, and pointing from other directions was not demonstrated. In the same study, the ambiguity of pointing directions was addressed using an unsupervised learning approach, although precise pointing directions were not considered. Furthermore, the methodology was limited to a tabletop scenario with a restricted viewing angle, reducing its applicability to broader, real-world environments.


\subsection{Perception using Human Pose Estimation Models}
Many advances in HRI rely on human pose estimation models, often referred to as \textit{Skeleton} models \cite{Lorentz2023, wang2023study}. These models utilize depth or RGB cameras to identify a set of key points representing the pose of the human body. Recent work by Medeiros et al. \cite{Medeiros2021} employed the OpenPose library \cite{Osokin2018} to extract a 2D skeleton model of a user from a single RGB camera. The spatial pose of the user's arm was derived from the 2D image through calibration based on fixed distances for a specific user. However, this approach is tailored to a single user, and the pointing accuracy remains unclear. The pointing direction was analytically extracted from the detected arm joints and used to guide a drone to one of several building windows. Prominent skeleton models, such as MediaPipe \cite{Lugaresi2019}, VNect \cite{mehta2017vnect}, and OpenPose \cite{OpenPose2019}, offer real-time 3D pose estimation using a single RGB camera. However, results presented in this work demonstrate that these models yield suboptimal performance for accurately predicting pointing direction. Additionally, the key points of the skeleton are provided relative to a predefined root rather than the camera, making it impossible to determine the arm's location with respect to a robot equipped with a camera. More recently, Nakamura et al. \cite{Nakamura2023} introduced a transformer-based model for pointing direction estimation. While innovative, this method is limited to indoor environments, relies on a skeleton model, and does not approximate finger location. Consequently, it is unsuitable for directing robots to desired targets.

\subsection{Alternative Body-Based Approaches for Pointing Estimation}
Finger data is challenging to extract in unstructured environments due to the small area it occupies within the image. Consequently, some studies focus on alternative or additional body features. A common approach involves determining the pointing direction based on the posture of the forearm \cite{Nickel2007,Kuramochi2021}. Lai et al. \cite{Lai2016}, T{\"o}lgyessy et al. \cite{Tlgyessy2017} and Hu et al. \cite{Hu2022} utilized a depth camera to identify the vector formed by the elbow and wrist, as extracted from a skeleton model \cite{Shotton2013}. This vector was then intersected with the floor to compute the desired target location for a mobile robot. Notably, Lai et al. \cite{Lai2016} detected pointing gestures by checking whether the arm's height fell within a predefined threshold relative to the shoulder. Another approach identified the positions of the head and hand using a depth camera, considering the vector connecting them as the pointing direction \cite{Azari2019}. In this method, the hand closest to the camera was heuristically selected as the pointing hand. In contrast, Shukla et al. \cite{Shukla2015} proposed a method that focused solely on the hand, independent of body posture. This model-free, probabilistic approach aimed to identify the pointed object from a predefined set. However, it relied on a depth camera and was limited to a controlled setting with a single viewing angle above a tabletop.

\subsection{Pointing Recognition via Wearable Devices}
While visual perception remains the dominant approach for pointing and gesture recognition, wearable devices represent an alternative method worth exploring. For example, two Inertial Measurement Units (IMUs) were positioned on a user's forearm to enable pointing on a presentation screen \cite{Sikeridis2017}. Frequently, Electro-Myography (EMG) is employed alongside IMUs. EMG involves sensing electrical muscle signals and mapping them to limb movements \cite{Khokhar2010}. While IMUs provide estimates of motion and posture, EMG enhances functionality by enabling hand gesture recognition. Combined use of EMG and IMUs has been applied for gesture recognition and detection of forearm pointing direction \cite{Haque2015}. Similarly, these technologies were integrated to control a drone navigating through obstacles \cite{DelPreto2020}. However, IMUs are prone to drift and require frequent recalibration to maintain accuracy \cite{Jirak2020}. Another approach utilized an off-the-shelf smartwatch to facilitate pointing and interaction with a large display \cite{Pietroszek2017}. Despite their versatility, wearable devices necessitate carrying additional hardware on the body, which hinders spontaneous or occasional pointing by arbitrary users, as discussed in the previous section.

\subsection{State-of-the-art Summary}

Table \ref{tb:sota} provides a summary of the state-of-the-art methods for pointing detection and estimation. The table categorizes various characteristics of prior approaches, which include the following: \textit{Recognition}: whether the method identifies the occurrence of pointing; \textit{Body posture}: whether the method requires additional body posture information beyond the arm; \textit{Skeleton}: whether the method employs a skeleton library for human pose estimation; \textit{IF/FA/EF}: whether the method observes the direction of the Index Finger (IF), Forearm (FA), or Eyes-Finger (EF) vector; \textit{Indoor/Outdoor}: whether the method has been demonstrated in both indoor and outdoor environments; \textit{RGB/Depth/Wearable}: the sensing technology used for pointing data acquisition; and \textit{Continuous/Discrete}: whether the method provides continuous pointing estimation or classifies pointing gestures from a discrete set of regions or objects. The table highlights the significant reliance of vision-based methods on depth cameras, which restricts their applicability to indoor environments. Furthermore, the table shows that forearm detection remains the dominant approach, while methods observing the index finger are less common, particularly with RGB cameras.

%% file: pre.tex
\begin{figure}
    \centering
    \includegraphics[width=0.7\linewidth]{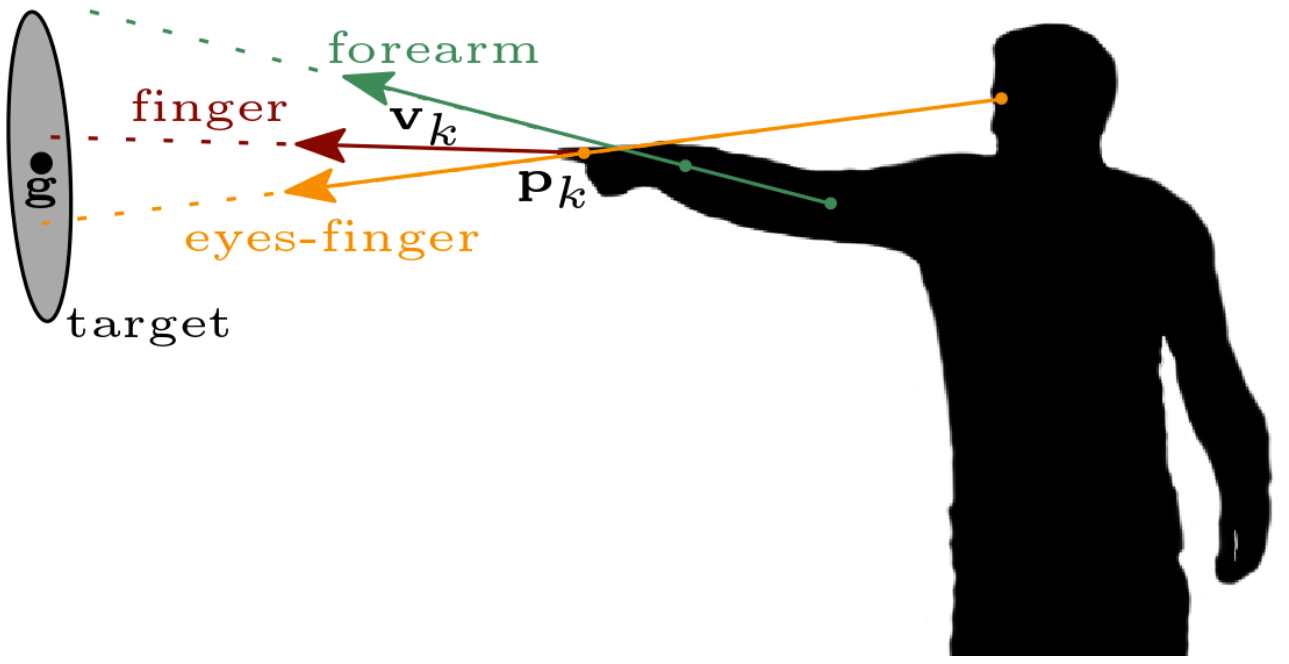}
    \caption{Three approaches for measuring pointing direction: forearm vector formed by the wrist and elbow points, index finger vector, and the vector connecting the user eyes and finger. Pointing error $e_k$ is calculated according to the minimal distance between the target center point $\ve{g}$ and the calculated direction vector $\ve{v}_k$ passing through some key-point $\ve{p}_k$ where $k=\{\texttt{FA},\texttt{IF},\texttt{EF}\}$.}
    \label{fig:finger_arm_eyes}
\end{figure}

As discussed above, prior work on pointing gestures has explored measuring either the forearm (FA) vector, formed by the wrist and elbow points \cite{Nickel2007,Kuramochi2021}, the direction of the index finger (IF) \cite{Jirak2020}, or the vector connecting the user’s eyes (root of the nose) to the finger (EF) \cite{Kehl2004, Nickel2007, Azari2019}. An illustration of these measurement types is provided in Figure \ref{fig:finger_arm_eyes}. However, to the best of the authors' knowledge, no comprehensive experimental comparison has been conducted to evaluate the accuracy of these three measurement approaches. In this section, we aim to analyze the accuracy of each measurement type in relation to the user’s intent and to determine whether observations of the IF alone are sufficient for the proposed application.

To evaluate the accuracy of each of the aforementioned measuring approaches, an experiment was designed and conducted with seven participants who were instructed to point toward a predefined target. Several reflective markers were positioned on designated eyeglasses, finger and forearm (wrist and elbow) of the user as well as on the target. Using a motion capture system, the target position along with poses of the user's index finger, forearm, and root of the nose were captured while pointing to the target. Users were instructed to point to the target in various positions with respect to the target and various body poses such as standing, sitting down and crouching. Also, the users were specifically asked to look at the target during pointing. For each measuring approach, the pointing direction $\ve{v}_k\in\mathbb{R}^3$ and a key-point $\ve{p}_k\in\mathbb{R}^3$ were calculated where $k=\{\texttt{FA},\texttt{IF},\texttt{EF}\}$. The key point is the position of the pointing finger for the IF and EF approaches, or the wrist for the FA approach. Then, the pointing error $e_k$ is the minimal distance between a line in direction $\ve{v}_k$ passing through $\ve{p}_k$ and the target center point $\ve{g}\in\mathbb{R}^3$, given by
\begin{equation}
    e_k=\frac{\|(\mathbf{g}-\mathbf{p}_k)\times\mathbf{v}_k\|}{\|\mathbf{v}_k\|}.
\end{equation}
Note that the experiment does not consider gaze but solely the position of the root of the nose as commonly done in pointing recognition work. Measuring gaze usually requires a camera fixed on the head of the user tracking pupil directions.

\begin{figure}
    \centering
    \includegraphics[width=\linewidth]{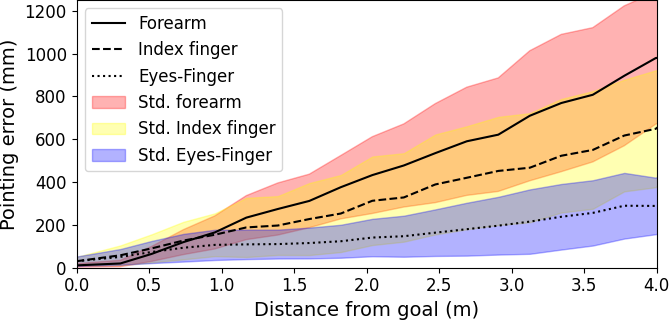}
    \caption{Pointing accuracy with regards to the distance of the user from the pointed target for three measurement approaches.}
    \label{fig:pointing_comparison}
\end{figure}

The comparative results of the pointing experiment for the three measuring approaches are presented in Figure \ref{fig:pointing_comparison}. The figure reports the mean and standard deviation pointing accuracies with respect to the distance from the target. The overall mean errors are 491.9~mm, 333.3~mm and 157.8~mm for the FA, IF and EF, respectively.  The results clearly indicate that FA measurement is the least accurate, as deviations at the wrist often result in angular discrepancies. In addition, EF measurement is the most accurate, while IF measurement exhibits slightly higher mean error values. Despite its superior accuracy, the EF approach has notable limitations. First, an average of 51\% of individuals are more likely to exhibit congruent gaze with the target while pointing \cite{Caruana2021}. The degree of variability between individuals is high and some may settle for an implicit gaze with no observable head shift. In addition, turning the head and gazing toward the target is performed for a short period of time of approximately 670 milliseconds \cite{Gomaa2020}. This is even more significant during cognitive overload where the user is performing various tasks such as during driving. Capturing the short time instant of user head-turn during pointing can be complex if it indeed occurs. Finally, having a pointing recognition model that depends both on the detection of finger and head limits the efficiency of the method in occluded environments where both must be visible. 

To conclude, while the experimental results in Figure \ref{fig:pointing_comparison} show that the EF method provides the highest accuracy on the test set, we have selected the IF measurement as the most practical approach for our model. This decision is driven by the IF method's superior robustness to common real-world challenges, such as head occlusion and subtle gestures that do not involve a deliberate head turn. The IF approach, therefore, better aligns with the goal of creating a robust, real-time pointing estimation system for unstructured environments.



%% file: Method.tex
\subsection{Problem Formulation}
\label{sec:problem_def}

We consider a scenario in which a user stands in front of a mobile robot equipped with a standard RGB camera mounted on it. It is assumed that the user is within the camera’s field of view. If the user is not visible, spatial voice localization can be used, for example, to attract the attention of the robot \cite{Gonzalez-Billandon2021}. However, such a problem is not in the scope of this work. Once the user exhibits a pointing gesture, the robot is required to recognize its occurrence, estimate direction, and move toward the pointed target. A scheme for this is illustrated in Figure \ref{fig:robot_scheme}. In this work, pointing is defined as the extension of the arm and index finger while the remaining fingers are flexed into the palm. Alternative gestures, such as pointing with multiple fingers or non-manual gestures like head nodding, could be explored in future studies \cite{Cooperrider2018}. 

\begin{figure}
    \centering
    \includegraphics[width=\linewidth]{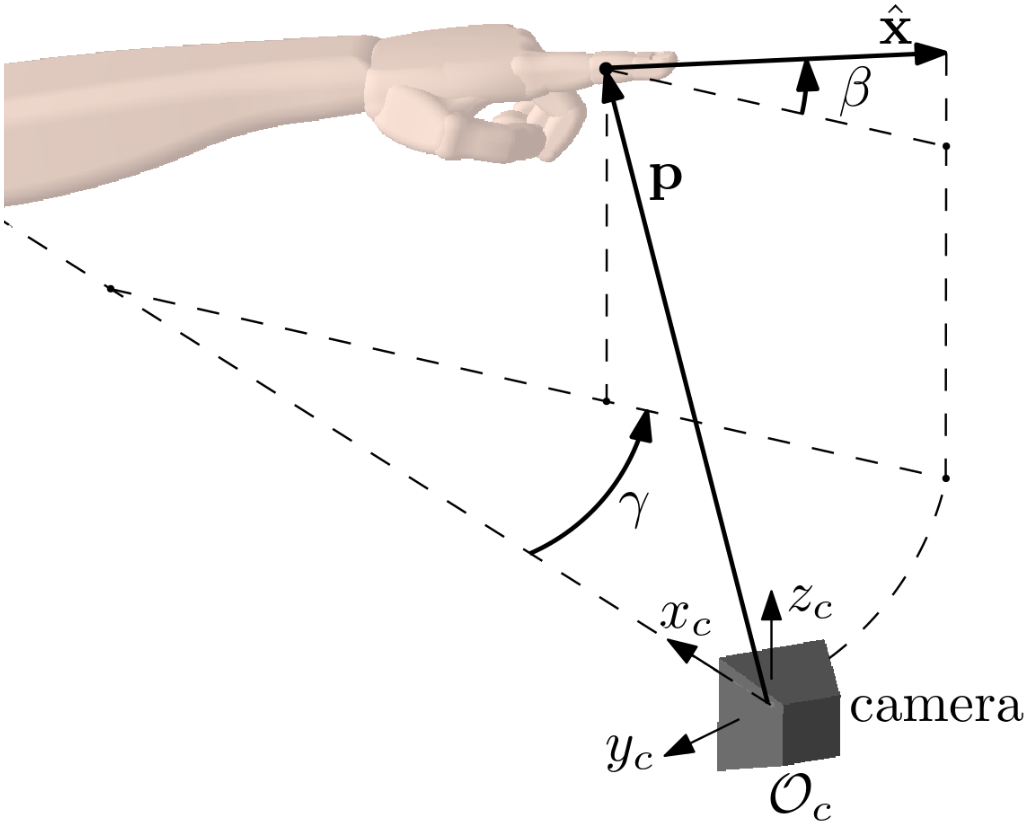}
    \caption{Illustration of the position $\ve{p}$ and direction $\hve{x}$ definitions of a pointing finger with respect to the coordinate frame of the camera $\mathcal{O}_c$. Direction can also be represented by angles $\beta$ and $\gamma$. }
    \label{fig:arm_pointing}
\end{figure}

To achieve the above goal and provide a pointing framework, we address two problems: 1) Derive a recognition model where the occurrence of a pointing gesture with the index finger is identified; and, 2) derive an estimation model for the pointing direction along with the position of the pointing finger. Images are continuously perceived by the robot in real-time. In parallel, a recognition model $h(I_t)\in\{0,1\}$ observes image $I_t$ at time $t$ for whether pointing is visible $h(I_t)=1$ or not $h(I_t)=0$. The recognition model acts as a trigger for direction estimation and motion to the target. The estimation model $\Gamma(I_t)$, in turn, is a regression problem aimed to map image $I_t$ to the following pointing parameters: index finger position $\ve{p}=(x_p,y_p,z_p)\in\mathbb{R}^3$ and pointing direction described by the pitch $\beta\in\mathbb{R}$ and yaw $\gamma\in\mathbb{R}$ angles. Pointing is, therefore, described by the feature vector $\ve{v}=(x_p,y_p,z_p,\beta,\gamma)$. These pointing parameters are defined relative to the coordinate frame $\mathcal{O}_c$ of the camera as seen in Figure \ref{fig:arm_pointing}. Yaw angle $\gamma$ is the rotation about the $z_c$ axis of $\mathcal{O}_c$. Pitch $\beta$, on the other hand, is the angle of the finger axis with the $x_c-y_c$ plane.


Without loss of generality, we consider the task of reaching a designated target goal. That is, the estimated pointing direction along with the position of the finger enables the robot to pinpoint the target location as illustrated in Figure \ref{fig:withGo1}. It is assumed that the user is pointing with only one arm, either right or left, while the second can do any other task. Motion planning will then be used to reach the target. In the general case and not in the scope of this work, some additional context may be required for the robot to reason about the target. For example, the user may instruct the robot to reach a door or a window. Similarly, the user can command the robot to bring a specific object. In both cases, the user will point in the corresponding direction and the robot will attempt to identify the target in the estimated direction. 

\begin{figure*}
    \centering
    \includegraphics[width=\linewidth]{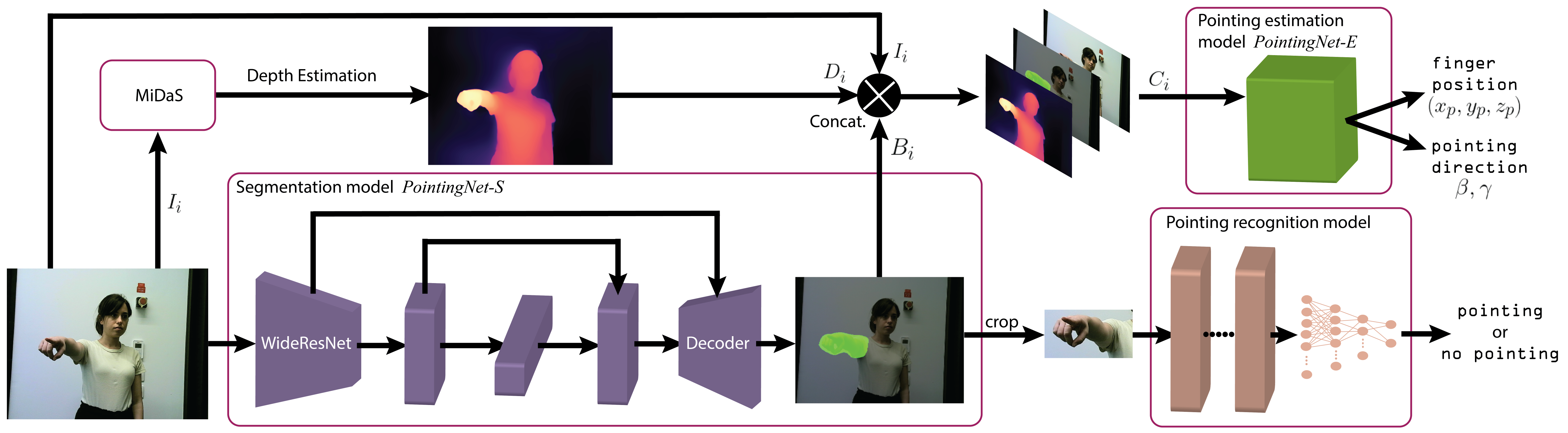}
    \caption{Flowchart of the proposed PointingNet model for recognizing the event of pointing, and estimating finger position and pointing direction. A segmentation model termed PointingNet-S is a main component in which a mask of a lifted arm is outputted. Then, the cropped arm image is passed through a CNN for classification of whether pointing is visible. If pointing is indeed identified, estimation model PointingNet-E is triggered where its input is the concatenation of the original image, the masked image and an estimated depth image generated by MiDaS. The output of the estimation model is the position of the finger and the direction of pointing, both with respect to the coordinate frame of the camera.}
    \label{fig:pointnet}
\end{figure*}

\subsection{Overview of Approach}
\label{sec:Overview}

To achieve the objectives outlined above, we propose the PointingNet model, depicted in Figure \ref{fig:pointnet}. The PointingNet framework comprises three main components: the PointingNet-S model for arm segmentation, a recognition model for identifying the occurrence of pointing gestures, and the PointingNet-E model for estimating the feature vector $\ve{v}$. In most scenarios, users will perform pointing gestures in unstructured environments containing diverse objects in the background. These objects may obscure the pointing arm, complicating the distinction between foreground and background elements. This challenge becomes more pronounced as the distance between the user and the camera increases, capturing additional irrelevant objects within the frame. Consequently, segmentation serves as a crucial initial step to isolate the pointing arm and minimize the influence of irrelevant visual data. To address this, we propose training a segmentation model, PointingNet-S, to automatically mask any lifted human arm within the frame, whether it is performing a pointing gesture or not.


The segmentation is further used for pointing recognition and feature vector estimation. Segmentation can improve recognition by providing a clear and focused representation of the objects or regions of interest within an image. This allows a classifier to concentrate on the most relevant features and ignore background or irrelevant information, leading to better accuracy and reduced over-fitting. Hence, segmentation is used to crop the region of the arm and provide input to a classification network. We note that, in a preliminary study, a segmentation model was tested for only pointing arms which would obviate the need for the recognition model. However, the model provided poor results with many false negatives. Hence, segmentation for any lifted arm followed by a recognition model provides the best results. 

The segmented mask is further utilized, along with estimated depth data extracted from the RGB image, as input to the PointingNet-E model for estimating the feature vector $\ve{v}$. We next describe the collection process for three datasets, followed by detailed descriptions of the segmentation, recognition, and estimation models. Finally, we explain how the estimated information is mapped into robot motion to direct it toward the pointed target.





\subsection{Data Collection}
\label{sec:data_collection}

\begin{figure}
    \centering
    \begin{tabular}{cc}
        \includegraphics[height=4.2cm]{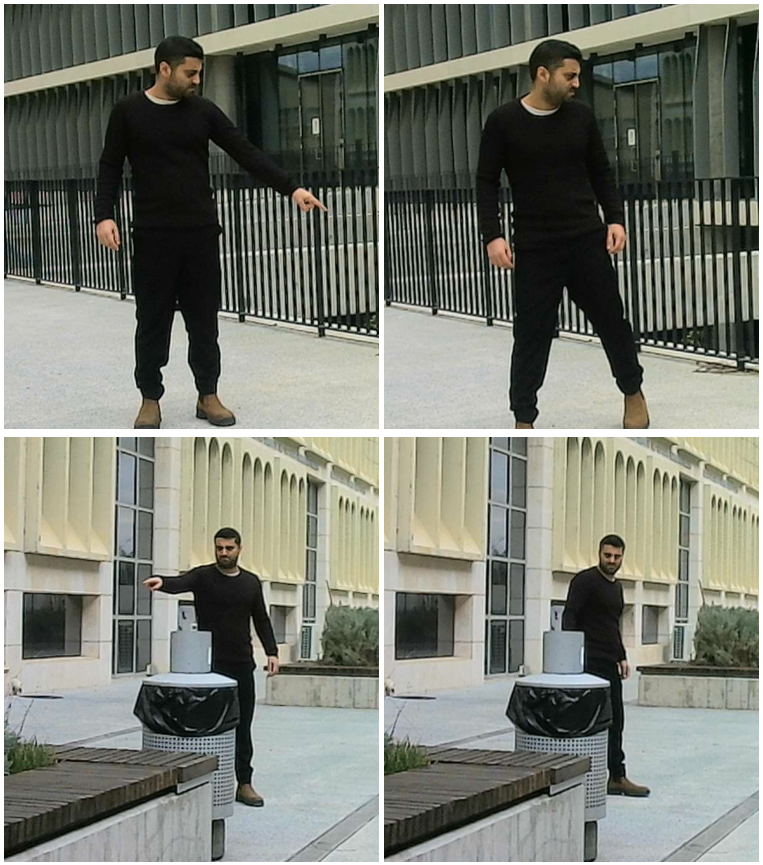} &  \includegraphics[height=4.2cm]{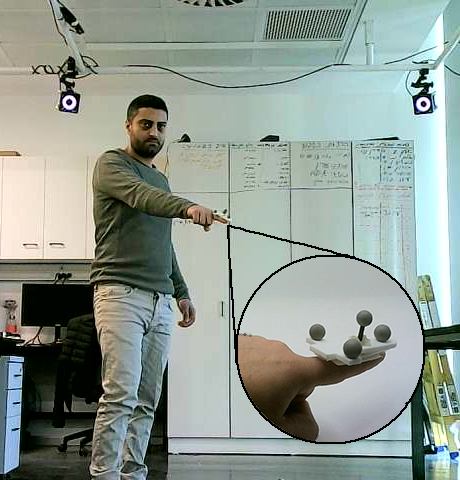}\\
        (a) & (b) 
    \end{tabular}
    \caption{Data collection for (a) recognition model by labeling images with pointing (left) or not pointing (right), and (b) estimation model by labeling images with a motion capture system observing a finger marker relative to a camera marker.}
    \label{fig:data_collection}
\end{figure}

Existing public datasets for pointing recognition fall short in key areas. They often lack annotations for the precise direction of the pointing finger, which is vital for accurate interpretation. Additionally, there's a notable absence of datasets featuring outdoor images, especially for ranges up to 5 meters, limiting training for real-world scenarios. To overcome these gaps, we've collected three new datasets specifically for training our model, all captured with a simple RGB camera. The first dataset $\mathcal{S}$ is collected for training model PointingNet-S. A set of $N_s$ images is collected where either the right or left arm is put in some arbitrary pose, not necessarily while pointing. Consequently, the dataset is in the form $\mathcal{S}=\{I_1,\ldots,I_{N_s}\}$ where $I_k$ is an image. Similarly, dataset $\mathcal{H}$ is collected for training the recognition model having $N_h$ images. Each image $I_i$ is labeled with $l_i\in\{0,1\}$ for whether pointing is visible ($l_i=1$) or not ($l_i=0$). Examples of labeled images are seen in Figure \ref{fig:data_collection}a. Thus, the resulting data is the set $\mathcal{H}=\{(I_1,l_1),\ldots,(I_{H_h},l_{N_h})\}$. 

The third dataset $\mathcal{P}$ includes data for training the PointingNet-E model. While the previous two datasets are collected in various indoor and outdoor environments, the third dataset requires a Motion Capture (MoCap) System and, therefore, is collected only indoors. Nevertheless, segmentation of the arm aims to allow the model to also perform well outdoors. Each image $I_j$ in $\mathcal{P}$ is taken by the RGB camera and labeled using the MoCap with feature vector $\ve{v}_j$. Vector $\ve{v}_j$ is obtained using a finger marker (with reflective markers as seen in Figure \ref{fig:data_collection}b) measured by the MoCap relative to a base marker on the RGB camera. The acquisition and labeling process yield dataset $\mathcal{P}=\{ (I_1,\ve{v}_1),\ldots,(I_{N_p},\ve{v}_{N_p}) \}$ of $N_p$ samples.

In all training datasets, data is collected by several users at various distances of the user from the camera in $[0.5,5]$ meters range, and in different locations within the image frame. Furthermore, we consider pointing with yaw angles in the range $\gamma\in[-125^\circ, 125^\circ]$. Angles not in the range will most likely not be visible to the robot and some intermediate pointing would have to be done to change its perspective. To generate a robust model, some images exhibit pointing where the user is occluded and most of the body is not visible except the pointing arm. Occlusion can occur by either an obstacle (example at the bottom of Figure \ref{fig:data_collection}a) or having a large portion of the user out of the image frame. In addition, pointing is performed while arbitrarily switching between right and left arms.

\begin{figure}
    \centering
     \begin{tabular}{cc}
        \includegraphics[width=0.45\linewidth]{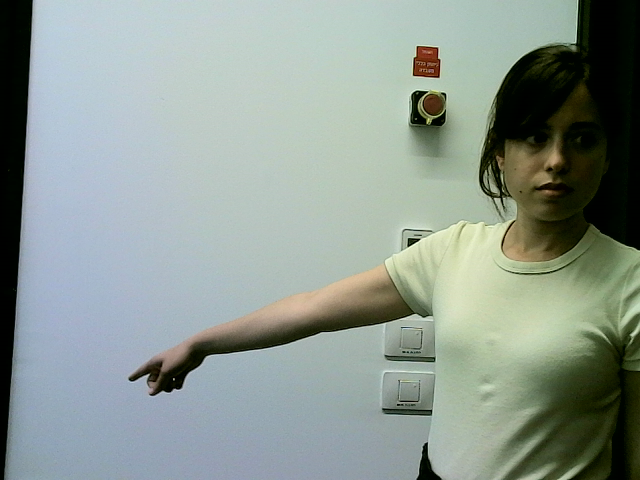} &  \includegraphics[width=0.45\linewidth]{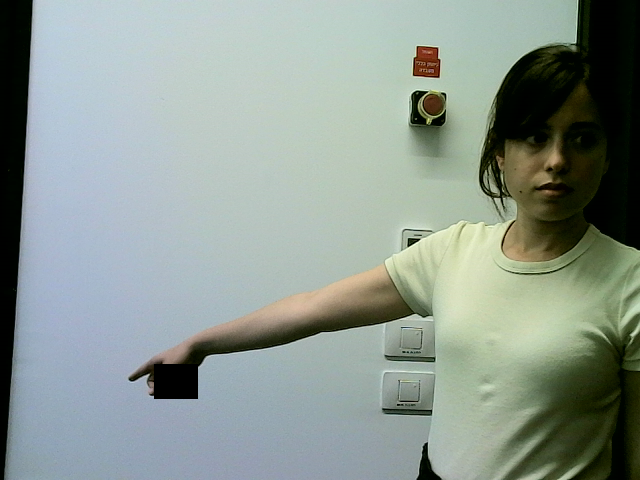}\\
        (a) & (b) \\
        \includegraphics[width=0.45\linewidth]{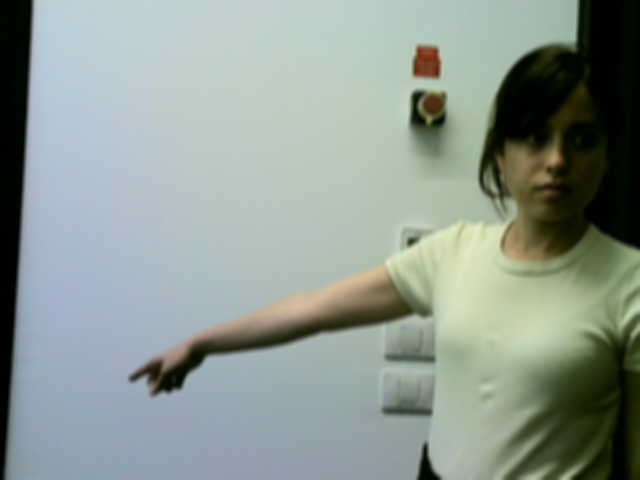} &  \includegraphics[width=0.45\linewidth]{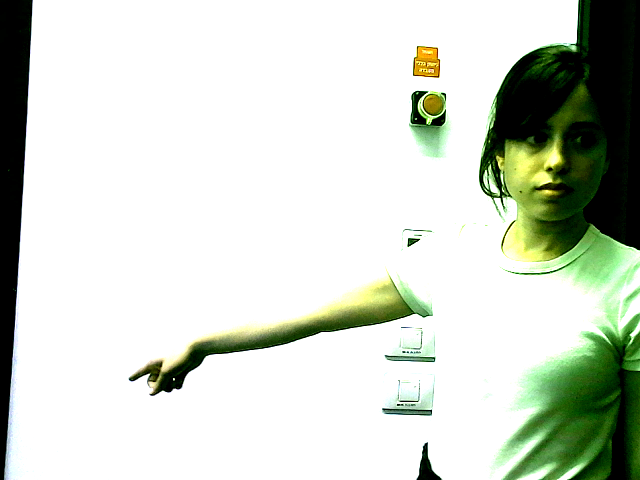}\\
        (c) & (d) \\
    \end{tabular}
    \caption{Augmentation examples of images in the training data: (a) original image, (b) random cut-out from image (black rectangle on hand), (c) image blur and (d) color jitter.}
    \label{fig:aumg}
\end{figure}

All images in the training dataset go through image augmentation. In addition to increasing the dataset, the augmentation encourages the model to recognize pointing in diverse and noisy environments, and also give partial information \cite{han2022}. Hence, the model will be able to perform well in changing environments which include lighting variations and occlusions. Applied augmentation techniques include color jitter, random Cutout and image blur. Augmentation examples of images in the training set are seen in Figure \ref{fig:aumg}.

\subsection{Pointing Segmentation}
\label{sec:pointing_segmentation}

A segmentation model breaks down an image into several segments of interest. Such a process enables reducing the complexity of the problem and assists in focusing on semantically meaningful regions for further image processing. In this part, we build a segmentation model that can segment the user's arm prior to identifying if pointing occurs. For that purpose, all images in dataset $\mathcal{S}$ are pre-processed and labeled by masking the corresponding user arm. The entire visible arm is masked including the hand and fingers. Hence, the pre-processed dataset is now $\tilde{\mathcal{S}}=\{(I_k,s_k)\}_{k=1}^{N_s}$ where $s_k$ is the set of masked pixels in image $I_k$. 

Using the collected data, a segmentation model termed PointingNet-S is proposed. First, a pre-trained Wide Residual Network (WideResNet) \cite{zagoruyko2016wide} is used with 101 layers to encode the images. WideResNet increases the width of the network, allowing it to capture complex features and patterns in the data. After encoding the image, the latent space representation is reshaped by a set of four convolutional operations, each having a different kernel. The first two kernels are odd-sized kernels of $11 \times 1$ and $1 \times 11$. An odd-sized kernel enables centering over the input data during convolution while having rotational symmetry. The symmetry can simplify the computation and make the network more robust to rotational variations in the input data. Two additional convolutions are Atrous with a rate of two and three. The Atrous convolution dilates the window size while incorporating gaps between data values so to increase the field of view without increasing the number of weights. 
Finally, the four convolutions are concatenated and fed into a standard decoder comprising 48 convolutional layers. In addition, skip connections are added between the encoder and decoder and between the convolutions to avoid overfitting, improve gradient flow, and encourage the reuse of features. The model is trained with dataset $\tilde{\mathcal{S}}$ and a Binary Cross-Entropy (BCE) with logits loss. BCE measures the difference between the predicted mask $\hat{s}_k$ and the ground truth mask $s_k$
\begin{equation}
\mathcal{L}_{\text{BCE}} = -\sum_{i=1}^{N} \sum_{j=1}^{m} \left[s_{i,j} \log \hat{s}_{i,j} + (1 - s_{i,j}) \log (1 - \hat{s}_{i,j})\right]
\end{equation}
where $m$ is the number of pixels in mask $s_k$. The segmentation model is illustrated in Figure \ref{fig:pointnet}. By leveraging the power of WideResNet, \textit{PointingNet-S} effectively captures complex image features, including fine-grained details such as fingertips. Atrous convolutions expand the receptive field without increasing computational cost, enhancing the model's ability to discern spatial relationships within images. Furthermore, skip connections facilitate efficient gradient flow and prevent overfitting, leading to improved model stability and accuracy. A customized convolution module and a tailored loss function further optimize the model's performance, enabling it to excel in challenging scenarios with multiple participants, occlusions, and varying viewpoints. In conclusion, we acquire a segmentation model $\hat{s}_i=S(I_i)$ that outputs the predicted set of pixels $\hat{s}_i$ masking any lifted arm in image $I_i$.

\subsection{Pointing Recognition}
\label{sec:pointing_recognition}

Object recognition refers to techniques or algorithms for recognizing objects within images. Neural networks, which are trained to recognize patterns in data, are efficient for such tasks and include techniques of classification, detection and segmentation \cite{krizhevsky2017imagenet, szegedy2015going}. In our pointing recognition problem, the task involves identifying whether the user seen in a query image exhibits a pointing gesture. To eliminate background noise and interference, the PointingNet-S model described above is used to detect any lifted arm in the image, crop out the rest of the image, and proceed to further classification. Therefore, dataset $\mathcal{H}$ is pre-processed to include only cropped images of lifted arms. That is, each image $I_i\in\mathcal{H}$ is passed through the segmentation model and cropped to the minimal rectangular area bounding the predicted label $\hat{s}_i=S(I_i)$. The pre-processed dataset $\tilde{\mathcal{H}}=\{(\tilde{I}_i,l_i)\}_{i=1}^{N_h}$ is acquired where $\tilde{I}_i$ is the cropped image of $I_i$.






Using $\tilde{\mathcal{H}}$, a recognition model $h(I_i)$ is trained for binary classification of images. Specifically, the model would determine whether an arm seen in an inputted cropped image exhibits pointing or not. Various classification models can be used to solve the pointing recognition task. Comparative analysis to be presented in the evaluation section shows that marginal performance differences exist between simple NN models (e.g., Fully-Connected NN or Convolutional-NN) to more complex pre-trained models (e.g., GoogLeNet \cite{szegedy2015going}). In addition, the former provides lower computational costs and is more suitable for real-time predictions. Therefore, we propose the use of a Convolutional NN (CNN) architecture with four convolutional layers. The output of the final convolutional layer is flattened and fed into a series of eight fully connected (FC) layers used for binary classification. The output of the final FC layer is passed through a Sigmoid activation function to obtain a predicted probability for whether the input image exhibits pointing. The CNN model and the ones used for comparison are trained using the BCE loss function.

\subsection{Pointing Estimation}
\label{sec:pointing_direction}

Once a pointing gesture has been recognized, the pointing feature vector $\ve{v}$ is to be estimated. As discussed in Section \ref{sec:introduction}, common approaches for human pose estimation are based on learning keypoints on the human body forming a skeletal representation (e.g., MediaPipe \cite{bazarevsky2020blazepose} and VNect \cite{mehta2017vnect}). As will be shown in the Evaluation Section, computing either forearm \cite{Medeiros2021, Lai2016} or index finger direction from these keypoints results in poor estimation accuracy. In addition, these keypoints are measured with respect to some arbitrary coordinate frame and not relative to the camera. Hence, state-of-the-art human pose estimation approaches cannot be used to extract the position of the pointing finger.

The prominent disadvantage of using regular RGB images is the lack of spatial information and, in particular, the user's 3D pose relative to the camera. Therefore, we incorporate the MiDaS Monocular Depth Estimation model \cite{Ranftl2020, Ranftl2021} which adds estimated depth information to the observed RGB image. MiDaS uses a single monocular camera to estimate the depth of objects in the scene with respect to the camera. Hence, each image $I_j\in\mathcal{P}$ is passed through the MiDaS model to generate a corresponding RGB depth image $D_j$. Furthermore, arm segmentation can focus the model on regions of interest. Hence, the predicted mask $\hat{s}_j=S(I_j)$ is used to generate a binary image $B_j$ (i.e., white and black for arm and non-arm regions, respectively). Finally, the images $I_j$, $D_j$ and $B_j$ are concatenated to a seven-channel image $C_j$. This configuration allows the observation of spatial relations, omission of irrelevant features, and reduction of the required data for training the pointing estimation model. While the full arm is segmented, the label for training PointingNet-E is the direction of the index finger. The index finger is included in the segmented arm. Also, the model receives additional information in the image, both in $I_j$ and $D_j$, including the standing pose of the human body with approximated depth information. Therefore, the model can learn the dependencies between body and arm postures and the direction of the index finger.

The proposed PointingNet-E for pointing estimation encodes the multi-channel image $C_j$ to acquire an estimated feature vector $\tve{v}_j=\Gamma(C_j)$. PointingNet-E is based on an eight-layers ConvNeXt encoder \cite{liu2022convnet}. The key innovation of ConvNeXt is its use of channel groups within the network to capture diverse features in input images efficiently. ConvNeXt can capture low-level and high-level features while maintaining a low computational cost by combining different types of convolutions within a block. We modify the ConvNeXt to include skip-connections. That is, the output of layer $i$ is passed to layer $i+1$ and also concatenated with the output of layer $i+2$ after down-sampling. These multiple pathways provide a form of redundancy that enables advanced information sharing and makes the network more robust to noisy or incomplete data. 

The output of the ConvNeXt encoder is further passed through three different Fully Connected Neural Network (FC-NN) blocks. Preliminary results have shown that it is significantly more accurate to estimate each feature in pointing vector $\ve{v}_i$ with a designated FC-NN rather than with a single FC-NN for all features. Hence, each of the three FC-NN blocks outputs either angle $\beta_i$, angle $\gamma_i$ or position $\ve{p}_i$ of pointing feature vector $\ve{v}_i$. All FC-NNs consist of four FC layers having Rectifier Linear Unit (ReLU) activation functions in between. However, only the FC-NNs of the angles have a Sigmoid activation function at the output. The model is trained to minimize a loss function summing the Root Mean Square Error (RMSE) and Mean Absolute Error (MAE) of the angles and position, respectively, over dataset $\mathcal{P}$. Hence, the loss function is defined by
\begin{equation}
    \mathcal{L} = \mathcal{L}_{pos} + \lambda \mathcal{L}_{dir}
\end{equation}
where $\mathcal{L}_{pos}$ is the RMSE loss for the position
\begin{equation}
\label{eq:Lpos}
\mathcal{L}_{pos} = \sqrt{\frac{1}{N_p} \sum_{i=1}^{N_p} \left\| \mathbf{p}_i - \tilde{\mathbf{p}}_i \right\|^2}
\end{equation}
and $\mathcal{L}_{dir}$ is the MAE loss for the direction angles
\begin{equation}
\label{eq:Ldir}
\mathcal{L}_{dir} = \frac{1}{N_p} \sum_{i=1}^{N_p} \left( \left| \beta_i - \tilde{\beta}_i \right| + \left| \gamma_i - \tilde{\gamma}_i \right| \right).
\end{equation}
The terms $\tilde{\mathbf{p}}_i$, $\tilde{\beta}_i$ and $\tilde{\gamma}_i$ are the predicted values of ${\mathbf{p}}_i$, ${\beta}_i$ and ${\gamma}_i$, respectively. Scalar $\lambda$ is a weighting factor to balance the contributions of the position and direction losses. In conclusion, given an image $I_j$ with exhibited pointing, depth estimation $D_j$ and binary arm mask $B_j$ are extracted so to estimate the feature vector $\tve{v}_j=\Gamma(C_j)$.

\subsection{Robot motion}
\label{sec:robot}

\begin{figure}
    \centering
    \includegraphics[width=\linewidth]{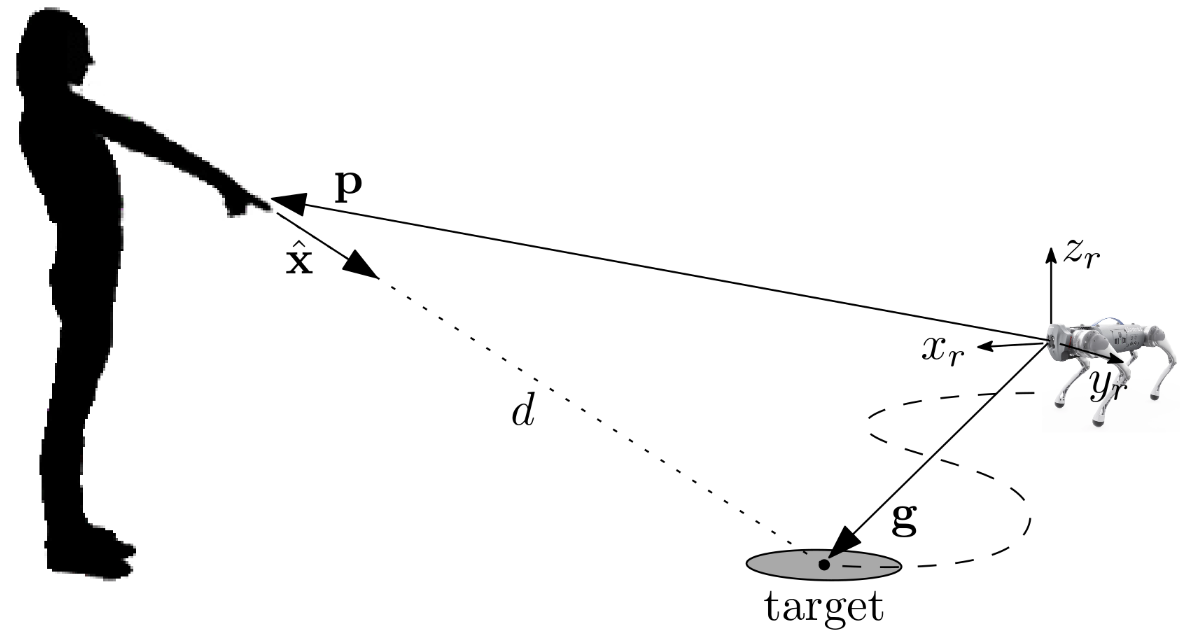} 
    \caption{Once the target $\ve{g}$ is computed, the robot can plan and control the motion to it (dashed curve). All vectors are expressed in the coordinate frame of the robot $\mathcal{O}_r$ after mapping from frame $\mathcal{O}_c$.}
    \label{fig:motion}
\end{figure}

In the idling state, the robot will loop through a stream of real-time images acquired from the camera. The images will pass through the pointing recognition model $h$. Once a pointing gesture is identified, the estimation model $\Gamma$ is triggered to output an estimated feature vector $\tve{v}$. Hence, the estimated position of the finger $\tve{p}^{(c)}$ and direction angles $\tilde{\beta},\tilde{\gamma}$ are now known. The superscript $(\cdot)^{(c)}$ indicates coordinates with respect to the frame $\mathcal{O}_c$ of the camera. The unit vector $\hve{x}^{(c)}$ in the direction of pointing is acquired by 
\begin{equation}
    \hve{x}^{(c)} = \begin{pmatrix}
\cos\tve{\gamma}\cos\tve{\beta}\\
\sin\tve{\gamma}\cos\tve{\beta}\\
-sin\tve{\beta} 
\end{pmatrix}.
\end{equation}
Given the homogeneous transformation matrix $A_{r,c}\in SE(3)$ mapping from $\mathcal{O}_c$ to the robot coordinate frame $\mathcal{O}_r$, finger position and pointing direction are expressed in $\mathcal{O}_r$ according to
\begin{equation}
\begin{pmatrix}
\hve{x}^{(r)}\\
1
\end{pmatrix}=A_{r,c}
\begin{pmatrix}
\hve{x}^{(c)}\\
1
\end{pmatrix}
\end{equation}
and
\begin{equation}
\begin{pmatrix}
\tve{p}^{(r)}\\
1
\end{pmatrix}=A_{r,c}
\begin{pmatrix}
\tve{p}^{(c)}\\
1
\end{pmatrix}.
\end{equation}
Furthermore, it is assumed that the environment of the robot is fully known such that distance $d$ from the finger to the floor surface along vector $\hve{x}^{(r)}$ can be estimated. For instance, if both user and robot are on a flat horizontal floor, distance $d$ can be calculated according to simple proportionality
\begin{equation}
    \label{eq:distance}
    d=-\frac{\tilde{p}_z^{(r)}+H}{\hat{x}_z^{(r)}}
\end{equation}
where $\tilde{p}_z^{(r)}$ and $\hat{x}_z^{(r)}$ are the $z$ components of $\tve{p}^{(c)}$ and $\hve{x}^{(r)}$, respectively, and $H$ is the height of the robot. Generally, the target location $\ve{g}$ in $\mathcal{O}_r$ for the robot to reach is acquired according to the resultant vector 
\begin{equation}
    \label{eq:target_g}
    \ve{g}=\tve{p}^{(r)}+\hve{x}^{(r)}d
\end{equation}
as seen in Figure \ref{fig:motion}. Once the target has been identified, the robot can walk directly to it if the environment is known to be obstacle-free. Alternatively, the robot can apply some motion planning algorithm \cite{lavalle2006} to plan a more complex path considering obstacles in the environment.

%% file: Evaluation.tex
This section tests and analyzes the proposed PointingNet model including analysis of arm segmentation, recognition model for classifying occurrences of pointing and direction estimation. We describe the collection of training and test data followed by an analysis of the model components. All model training and evaluations were performed on a Linux Ubuntu (18.04 LTS) machine with Intel Xeon Gold 6230R (20 Cores of 2.1GHz) CPU and four NVIDIA GeForce RTX 2080TI GPU (11GB of RAM). Neural network models were trained and tested using PyTorch. All models were trained using the ADAM optimizer with adaptive learning rate. Videos of the data collection, experiments and demonstrations can be seen in the supplementary material.


\subsection{Database and Descriptors}

As discussed in Section \ref{sec:data_collection}, three datasets $\mathcal{S}$, $\mathcal{H}$ and $\mathcal{P}$ were collected for arm segmentation, pointing recognition and pointing direction estimation, respectively. For the collection of these datasets, 12 participants were recruited, with nine contributing to the training data and three to the test data. Participants and recorded environments in the test set were not included in the collection of training data. Participants include four females and eight males with maximum and minimum heights of 1.56 m and 1.91 m, respectively. 
All image data was collected with a simple web camera yielding images of size $480\times640$. However, to enable robust usage with any camera, the images were reduced to resolution $288\times384$. To enhance the labeling process of the information, the acquired data for segmentation was obtained at a resolution of $1920\times 1080$ and later reduced. This allows for a more comprehensive visualization of objects or regions of interest and facilitates the identification and annotation of such elements. 

Both datasets $\mathcal{S}$ and $\mathcal{H}$ were collected by the participants in indoor and outdoor environments as the MoCap system is not required. Participants stood in various locations in front of the camera and up to five meters from it. Each participant was asked to randomly point at any desired direction, with the only constraint being that they point within the range of $\gamma\in[-125^\circ, 125^\circ]$. Participants arbitrarily switched between their right and left arms. Also, participants arbitrarily chose to have short or long sleeves during recording. The collection yielded $N_s=17,698$ and $1,800$ train and test samples, respectively, for dataset $\mathcal{S}$. In addition to many images of pointing, the dataset includes images of lifted arms in various tasks such as waving or holding an object. To allow for robust segmentation models, some images included scenarios without any participant or with features resembling a human arm (e.g., unused coat or shirt). These images were given a blank mask (i.e., no segmentation). To generate the masked dataset $\tilde{\mathcal{S}}$, images in $\mathcal{S}$ were pre-processed and labeled by masking the corresponding user arm with the \textit{V7 auto-annotate} tool. 

Dataset $\mathcal{H}$ for classification includes $N_h=20,000$ training samples and $2,000$ test samples. Half of $\mathcal{H}$ is of participants exhibiting pointing in the images and labeled $l_i=1$. Images of the other half, labeled $l_i=0$, include participants doing various tasks while standing, walking, sitting or crouching but without pointing. The third data set $\mathcal{P}$ includes $N_p=186,300$ and $17,458$ train and test samples, respectively, of images labeled with pointing feature vectors. Feature vector $\ve{v}_i$ was measured by a MoCap system with eight OptiTrack Prime 41 cameras as discussed in Section \ref{sec:data_collection} and demonstrated in Figure \ref{fig:data_collection}b. Since a MoCap was used, the collection was only performed indoors. 





\subsection{PointingNet-S evaluation}

As discussed in Section \ref{sec:pointing_segmentation}, segmentation focuses the recognition and estimation models on the user's arm. PointingNet-S is a designated model for the segmentation of lifted arms in images. We compare PointingNet-S to various state-of-the-art segmentation models including Autoencoder, U-Net, U-Net++, Feature Pyramid Network (FPN), ResNet-101, DenseNet and DeepLabV3+. The baseline is a simple Autoencoder in which the original image is encoded and reconstructed to the masked variant \cite{Karimpouli2019}. U-Net is a widely used image segmentation model for biomedical applications where the image is contracted and expanded in a U-shaped path of convolutional layers \cite{ronneberger2015u}. U-Net++ is an extension of the U-Net architecture with additional skip-connections to improve segmentation accuracy \cite{zhou2019unet++}. Similarly, FPN uses a pyramid-shaped feature hierarchy to extract features at different scales for object segmentation \cite{lin2017feature}. ResNet-101 \cite{he2016deep} and DenseNet-201 \cite{huang2017densely} are deep learning models that use residual connections and densely connected blocks, respectively, to improve accuracy and reduce overfitting. The DeepLabV3+ model uses dilated convolution and an encoder-decoder structure with a ResNet-101 backbone to preserve fine details \cite{chen2018encoder}. 
These models are commonly used to identify and segment various objects in images in different applications. They provide different trade-offs between accuracy and computational efficiency. Therefore, the comparison is important because the performance of the models varies depending on the specific segmentation task and dataset.

\begin{figure}
    \centering
     \includegraphics[width=\linewidth]{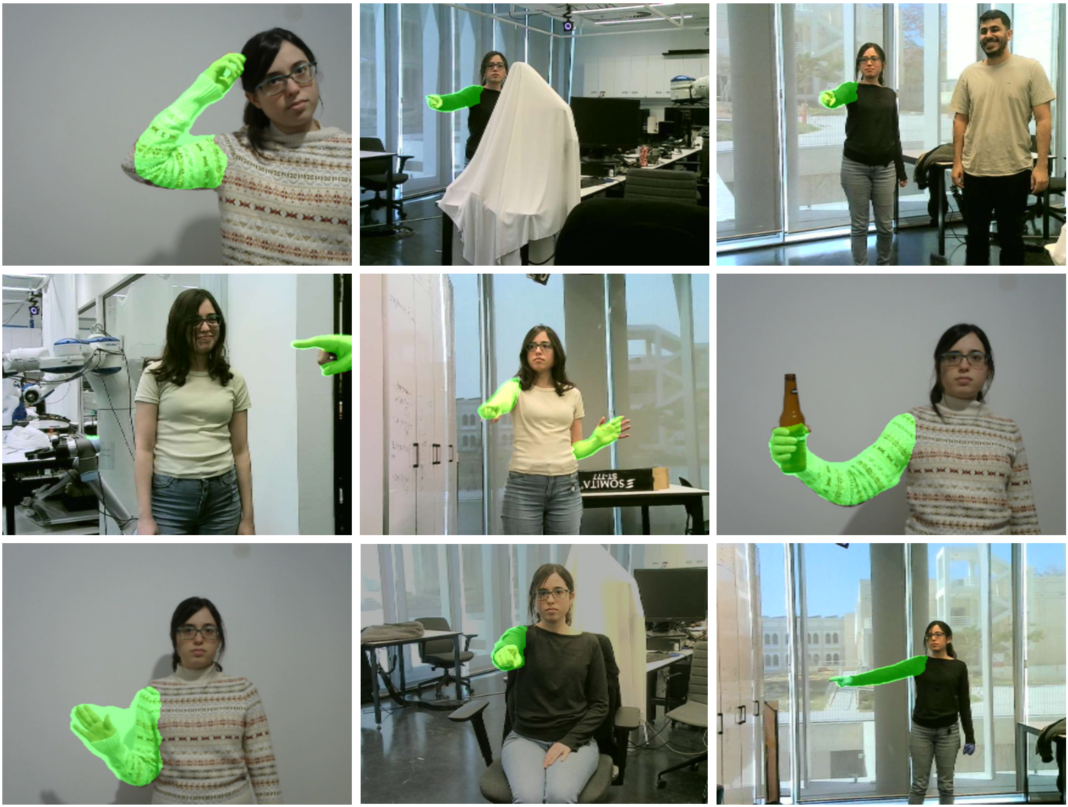} 
    \caption{Arm segmentation examples outputted by PointingNet-S in various scenarios of multiple participants, occlusions, dual-arm lift, sitting down and holding objects. The segmentation is performed for any lifted arm regardless of pointing or not. Classification later determines if pointing indeed occurs.}
    \label{fig:segmentations}
\end{figure}

All evaluated models are trained with dataset $\tilde{\mathcal{S}}$ and evaluated on the independent test set. The performances of all models are evaluated with the Intersection over Union (IoU). IoU calculates the percentage ratio between overlapping and unified areas of the ground-truth and predicted masks. The hyper-parameters of \textit{PointingNet-S} were optimized to provide the highest IoU.
Table \ref{tb:Segmentation} reports the comparative IoU results between all methods. While some models achieved adequate accuracy, PointingNet-S is shown to provide superior results by a large margin. Figure \ref{fig:segmentations} shows examples of predicted segmentation of the arms in various scenarios. PointingNet-S is able to segment lifted arms in complex scenarios including multiple users in the frame, occlusions, and user pointing while the body is out-of-frame. The proposed \textit{PointingNet-S} architecture effectively isolates complex image features associated with pointing gestures, including arm and hand configurations, while suppressing irrelevant background and foreground elements. The model's ability to segment the arm region demonstrates potential for broader applications in gesture recognition and human pose estimation, where focusing on relevant body parts can improve accuracy and efficiency in challenging environments.

\begin{table}[]
\centering
\caption{Arm Segmentation Evaluation Results}
\label{tb:Segmentation}
\begin{tabular}{lcc}\toprule
        Models  &~~~~~~~~~~~~~~~~~~& IoU  \\\midrule
        Autoencoder && 66.9\%  \\
        U-Net && 60.1\%  \\
        U-Net++ && 86.6\% \\
        FPN && 69.2\%  \\
        ResNet-101 && 78.5\% \\
        DenseNet-201 && 82.9\% \\
        DeepLabV3+ && 86.1\% \\
        PointingNet-S && \cellcolor[HTML]{C0C0C0} 94.3\%  \\\bottomrule
\end{tabular}
\end{table}

\subsection{Pointing Recognition Evaluation}

Once segmentation of the arm has been acquired, the image is cropped to include only the arm. As discussed in Section \ref{sec:pointing_recognition}, the cropped image is passed through the classification model to determine if pointing occurs. In our proposed approach, a CNN network is trained to perform the classification. We compare the model with various other models including ResNet-18, Visual Geometry Group (VGG) and GoogLeNet. VGG is a popular model for object recognition and segmentation that uses a deep convolutional architecture \cite{simonyan2014very}. Similarly, GoogLeNet is a deep CNN having a unique inception module that incorporates multiple filters in parallel \cite{szegedy2015going}. In addition, we compare classification with and without segmentation of the arm. When not including segmentation, the entire image is passed through the model.

\begin{figure}
    \centering
     \includegraphics[width=0.8\linewidth]{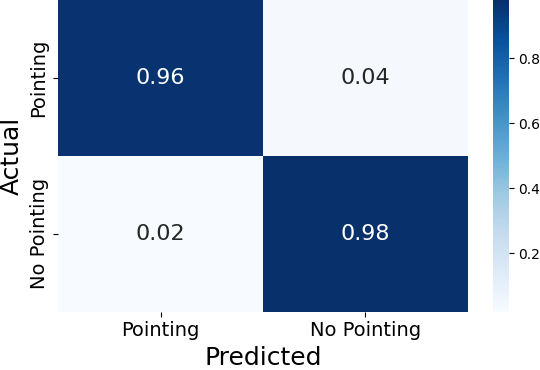}  \\
    \caption{A confusion matrix illustrating the performance of the CNN classifier in recognizing the occurrence of pointing gestures.}
    \label{fig:cm}
\end{figure}

\begin{figure}
    \centering
    \begin{tabular}{c}
         \includegraphics[width=0.7\linewidth]{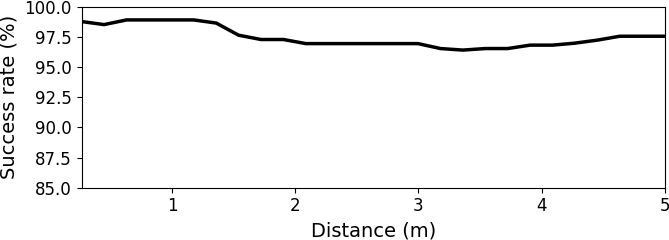}  \\
         \includegraphics[width=0.7\linewidth]{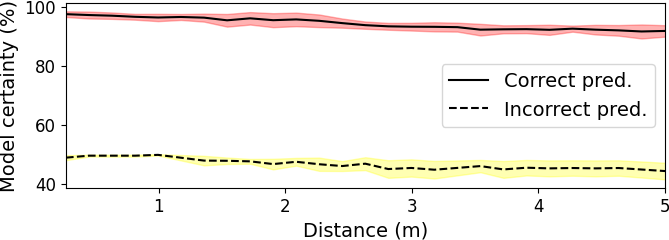} 
    \end{tabular}
    \caption{Classification (top) success rate and (bottom) certainty with regards to the distance of the user's arm from the camera.}
    \label{fig:error_distance1}
\end{figure}

Table \ref{tb:Classification} presents classification accuracy for all models, both with and without segmentation. The results underscore the importance of segmentation, as models without this preprocessing step exhibit significantly lower performance. While there are marginal improvements in accuracy when transitioning from simpler models, such as the CNN, to more complex architectures, the computational overhead associated with these complex models may outweigh the benefits. Consequently, the proposed CNN model is selected for subsequent experiments. Figure \ref{fig:cm} shows the confusion matrix for the CNN model's performance on the test set, with an accuracy of 97.2\%, precision of 97.9\%, and recall of 96.5\% for recognizing the occurrence of a pointing gesture. These metrics indicate that the model is both highly accurate in identifying pointing gestures and reliable in avoiding false positives. Figure \ref{fig:error_distance1} illustrates the relationship between classification accuracy, certainty, and user distance from the camera. The model demonstrates consistent high accuracy and certainty for correct classifications across varying distances. Conversely, incorrect classifications exhibit low certainty, suggesting a potential filtering mechanism to discard unreliable predictions.


\begin{table}[]
\centering
\caption{Pointing Recognition Evaluation Results}
\label{tb:Classification}
\begin{tabular}{lcccc}\toprule
        \multirow{2}{*}{Models}  &~~~~~~& \multicolumn{3}{c}{Segmentation} \\
        && w/o  &~~~~~~& w/ \\\midrule
        CNN && 66.7\% && 97.2\% \\
        ResNet-18 && 74.6\% && 98.1\% \\
        VGG-19 && 75.1\% && 98.5\% \\
        GoogLeNet && 78.9\% && 98.7\% \\
\bottomrule
\end{tabular}
\end{table}

\subsection{PointingNet-E Evaluation}

Once pointing is identified, PointingNet-E is triggered. In this section, we evaluate the performance of the model on the test data and in real-time predictions. PointingNet-E model was trained with dataset $\mathcal{P}$ as discussed in Section \ref{sec:pointing_direction}. Accuracy of the model over the test data is evaluated using RMSE in \eqref{eq:Lpos} for the finger position and MAE in \eqref{eq:Ldir} for yaw and pitch angles, all with respect to the measured ground-truth. The hyper-parameters of \textit{PointingNet-E} were optimized to provide the lowest loss value over the test data. 

\begin{table}[]
\centering
\caption{Pointing direction estimation results}
\label{tb:pointing_estimation}
\begin{adjustbox}{width=\linewidth}
\begin{tabular}{lcccc}\toprule
            && Position   & Yaw            & Pitch           \\
            && RMSE (mm)  & MAE ($^\circ$) & MAE ($^\circ$)  \\ \midrule
\multirow{2}{*}{MediaPipe} & (index finger) & - & 26.1$\pm$12.2 & 17.3$\pm$16.0  \\
 & (forearm) & -              & 29.6$\pm$10.7  & 15.8$\pm$12.5  \\
VNect    & (forearm) & - & 46.8$\pm$26.3   & 30.5$\pm$17.4  \\
OpenPose     & (forearm) & -     & 33.7$\pm$26.3   & 16.6$\pm$10.6 \\\midrule

CNN && 241.1$\pm$8.1  & 41.2$\pm$8.3   & 44.3$\pm$21.2 \\
VGG-19    && 95.4$\pm$4.7   & 29.1$\pm$7.8   & 40.4$\pm$19.7 \\
DenseNet	&& 162.8$\pm$79.4 & 21.9$\pm$10.1  & 13.5$\pm$4.5  \\
PointingNet-E & $I_i$ only      & 216.5$\pm$55.1 & 38.2$\pm$6.7  & 17.9$\pm$7.7 \\
PointingNet-E & $I_i$ \& $D_i$  & 94.3$\pm$4.18        & 7.17$\pm$6.1 & 3.43$\pm$2.4 \\
PointingNet-E & $I_i$ \& $B_i$  & 77.1$\pm$84.7  & 8.4$\pm$3.6 & 7.8$\pm$6.3 \\
\rowcolor{Gray}
PointingNet-E & $C_i$           & 61.3$\pm$30    & 1.4$\pm$1.3 & 0.61$\pm$0.63 \\ 
 \bottomrule
\end{tabular}
\end{adjustbox}
\end{table}

Table \ref{tb:pointing_estimation} presents an overall comparison of the proposed approach with various known models. First, we evaluate the pointing accuracy with state-of-the-art skeleton-based human pose estimation models including MediaPipe \cite{Lugaresi2019}, VNect \cite{mehta2017vnect} and OpenPose \cite{Osokin2018}. These models are pre-trained and used as provided through open-source distribution. While VNect and OpenPose offer spatial pose estimation of only the forearm, MediaPipe can provide spatial estimation of the finger as well. Hence, pointing is evaluated solely using forearm keypoints for the former two while using also the index finger for the latter. Human pose estimation models are not explicitly designed for pointing tasks and, therefore, exhibit significant limitations in accurately determining pointing direction. Additionally, the fingertip position coordinates estimated by these models are often relative to some unknown reference frame, hindering their direct application in camera-centric coordinate systems required for robot guidance. Consequently, such skeleton-based human pose estimation models are deemed unsuitable for precise pointing estimation and robot directives.


\begin{figure}
    \centering
    \begin{tabular}{c}
         \includegraphics[width=\linewidth]{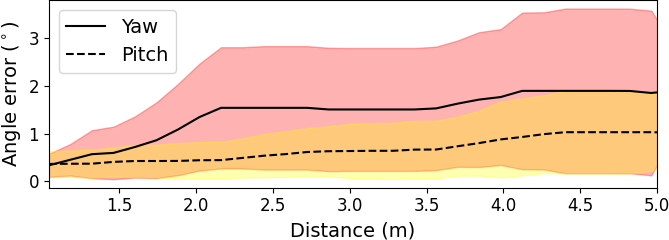}  \\
         \includegraphics[width=\linewidth]{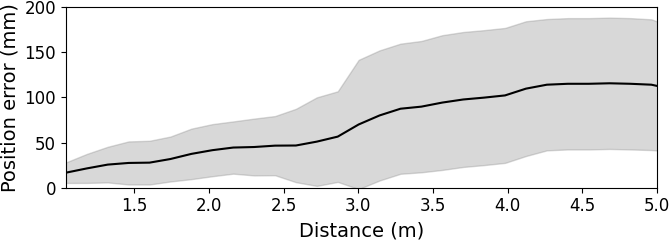} 
    \end{tabular}
    \caption{Prediction errors of (top) yaw and pitch angles and (bottom) position with regards to the distance of the user's arm from the camera.}
    \label{fig:error_distance2}
\end{figure}

\begin{figure}
    \centering
         \includegraphics[width=0.6\linewidth]{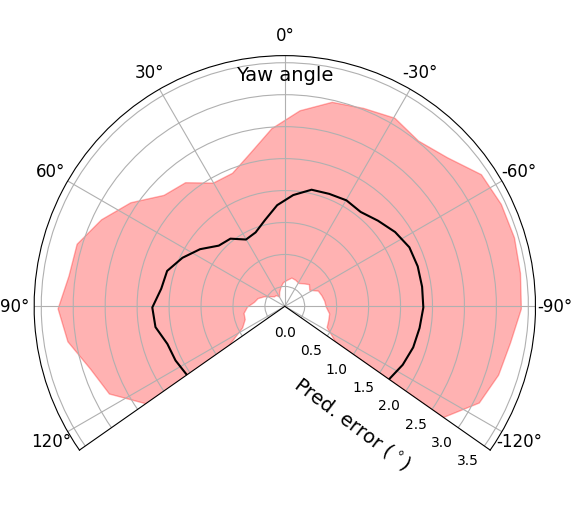}  \\
    \caption{Prediction error distribution of the yaw $\gamma$ angle.}
    \label{fig:yaw_polar}
\end{figure}

\begin{figure}
    \centering
         \includegraphics[width=0.6\linewidth]{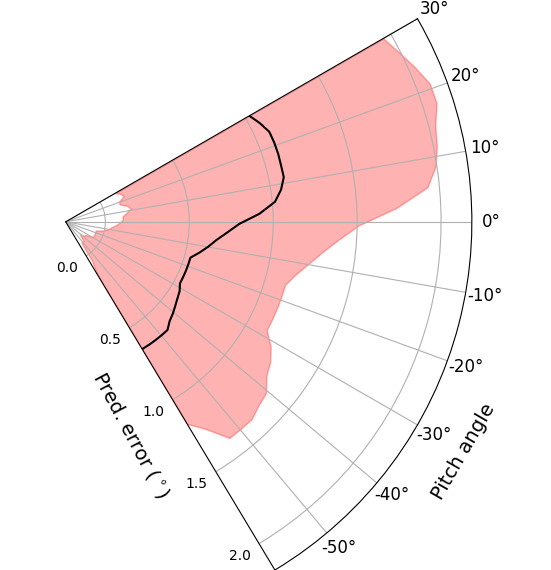} 
    \caption{Prediction error distribution of the pitch $\beta$ angle.}
    \label{fig:pitch_polar}
\end{figure}

The proposed model is next compared to various deep NN models including a simple CNN, VGG-19 and DenseNet. The input for these models is only the original image $I_i$. Furthermore, PointingNet-E is benchmarked across different inputs: only the original image $I_i$; a concatenation of $I_i$ with the MiDaS depth approximation $D_i$; a concatenation of $I_i$ with arm mask $B_i$; and, the concatenation $C_i$ of all three $I_i$, $B_i$ and $D_i$. The results in Table \ref{tb:pointing_estimation} show poor accuracy for CNN, VGG-19 and DenseNet as well as for PointingNet with only $I_i$, both in position and direction. However, including $B_i$ or $D_i$ to $I_i$ significantly reduces estimation errors. These results, therefore, highlight the importance of each of the three components in predicting the position and direction of pointing gestures. Consequently, having all three images $I_i$, $B_i$ and $D_i$ provides significantly lower position and direction errors (less than 2$^\circ$ in average). Figure \ref{fig:error_distance2} shows PointingNet-E estimation accuracy with regard to the distance of the user from the camera. The results show slight error increase with distance while still being fairly accurate. In addition, Figures \ref{fig:yaw_polar} and \ref{fig:pitch_polar} show the prediction errors of the yaw and pitch angles, respectively, with respect to the angles themselves. The plots show relatively uniform error distribution across the learned range. These results validate the effectiveness and feasibility of PointingNet-E for accurate estimation of pointing. 

\begin{figure}
    \centering
         \includegraphics[width=\linewidth]{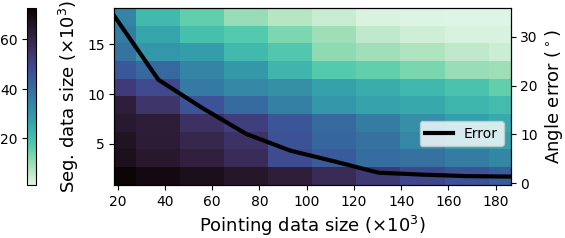} 
    \caption{Heatmap of the yaw angle error ($^\circ$) with regards to the sizes of $\mathcal{P}$ and $\mathcal{S}$ to train PointingNet-E and PointingNet-S, respectively. The black curve (with the right $y$-axis) shows the error of the yaw angle estimation with PointingNet-E when using the fully trained PointingNet-S. }
    \label{fig:yaw_acc}
\end{figure}

\begin{figure}
    \centering
         \includegraphics[width=\linewidth]{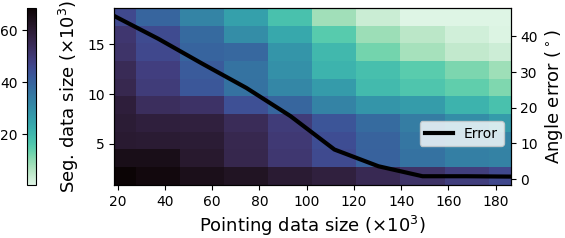} 
    \caption{Heatmap of the pitch angle error ($^\circ$) with regards to the sizes of $\mathcal{P}$ and $\mathcal{S}$ to train PointingNet-E and PointingNet-S, respectively. The black curve (with the right $y$-axis) shows the error of the pitch angle estimation with PointingNet-E when using the fully trained PointingNet-S. }
    \label{fig:pitch_acc}
\end{figure}

\begin{figure}
    \centering
         \includegraphics[width=\linewidth]{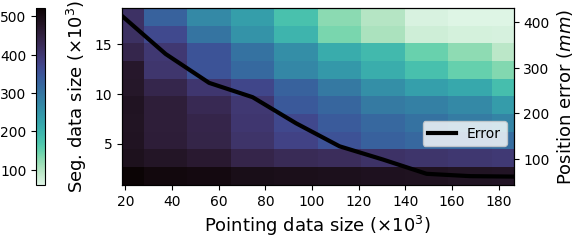} 
    \caption{Heatmap of the position error ($mm$) with regards to the sizes of $\mathcal{P}$ and $\mathcal{S}$ to train PointingNet-E and PointingNet-S, respectively. The black curve (with the right $y$-axis) shows the error of the position estimation with PointingNet-E when using the fully trained PointingNet-S. }
    \label{fig:pos_acc}
\end{figure}

Heatmaps of yaw, pitch and position estimation errors with respect to data sizes can be seen in Figures \ref{fig:yaw_acc}-\ref{fig:pos_acc}. The errors are calculated over the test data while increasing the number of points $N_p$ in $\mathcal{P}$ for PointingNet-E and the number of points $N_s$ in $\mathcal{S}$ for PointingNet-S. For each cell in the heatmap, PointingNet-S was trained with $N_s$ training data and then used to segment $N_p$ data for training PointingNet-E. The figures also show curves for the error during the increase of training data for PointingNet-E while having PointingNet-S trained with all available data. The accuracy increases with the addition of pointing examples. In addition, the results show the importance of a sufficiently trained segmentation model which also emphasize the contribution of arm segmentation to the pointing estimation accuracy.

\begin{figure}
    \centering
     \includegraphics[width=\linewidth]{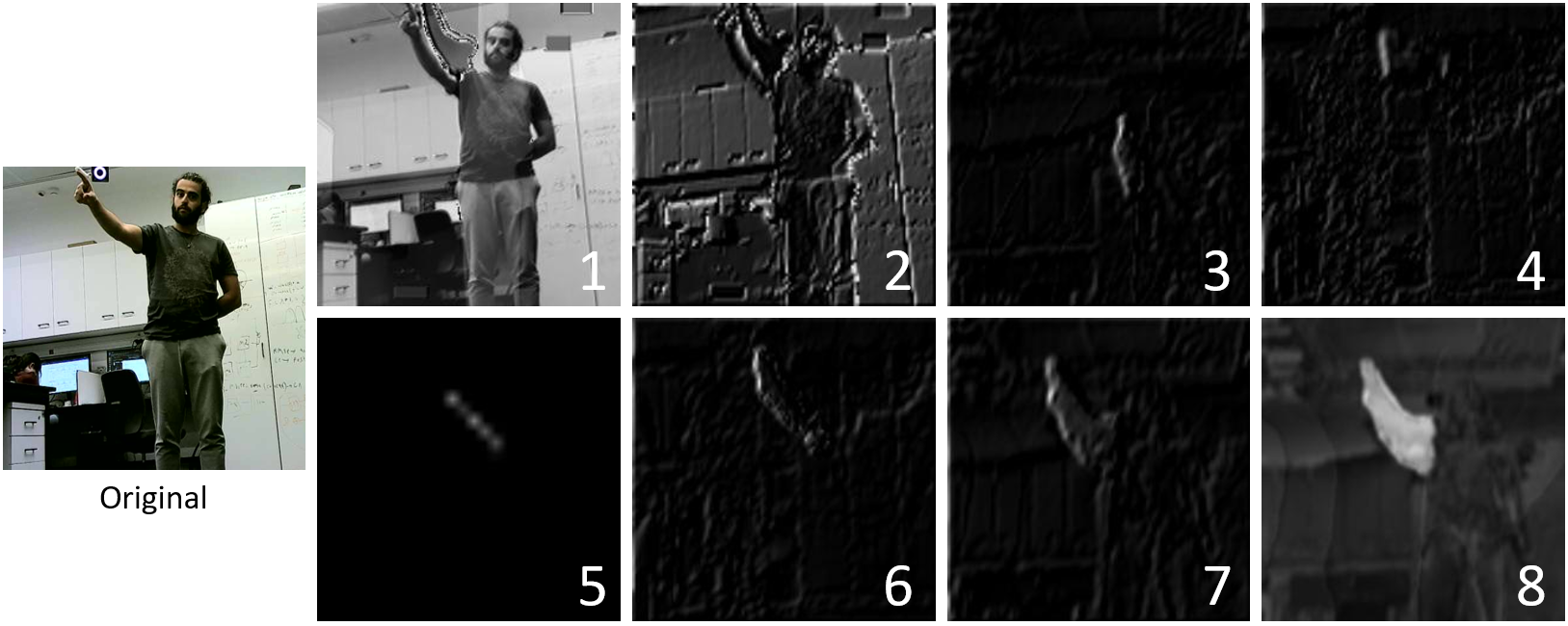} 
    \caption{Original image (left) of a pointing participant and the outputted images from the first (1) to last (2) layers of the ConvNeXt in PointingNet-E.}
    \label{fig:features}
\end{figure}

Figure \ref{fig:features} shows an example of one pointing test image inputted to PointingNet-E and the images extracted from its layers. The initial layers of the modified ConvNeXt predominantly detect the edges of the image with some focus on the arm. 
In the following layers, there is a stronger emphasis on the region of interest, i.e., the arm, compared to the background areas. The output of the fifth layer highlights the pose of the arm. In the final layers, the model prioritizes the finger and the entire arm region receiving the highest intensity. While it is hard to reason about the exact features learned by the model, visualization of the learned features in each layer of PointingNet-E provides implications that the model learns to distinguish between the different elements of the image and focuses on arm and finger posture. Presumably, PointingNet-E learns to predict finger position and direction by observing the index finger, but based also on learning patterns of arm postures. 






\begin{figure}
    \centering
     \includegraphics[width=\linewidth]{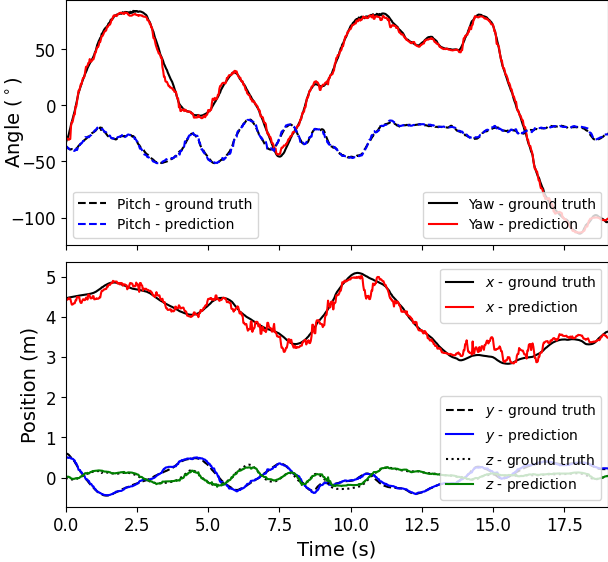} 
    \caption{Real-time prediction of pointing parameters including (top) yaw and pitch angles and (bottom) position of the user's finger.}
    \label{fig:session_example}
\end{figure}

Next, the ability of the model to provide pointing position and direction estimation in real-time is demonstrated. Here, the test participant moved the arm fairly slowly while changing pointing directions. The camera acquired images at a frequency of 38~Hz. Prediction accuracy matches the ones reported in Table \ref{tb:pointing_estimation}. Furthermore, Figure \ref{fig:session_example} shows an example of one real-time prediction session.  The mean yaw and pitch error angles along the session are 1.78$^\circ$ and 0.8$^\circ$, respectively. Similarly, the position errors along the $x$-, $y$- and $z$-directions are 102.7~mm, 20.66~mm and 21.83~mm, respectively. PointingNet-E predictions closely match the ground-truth with slightly larger errors along the $x$-axis. The experimental results conclusively demonstrate the model's capacity to accurately predict pointing direction and position in real-time, fulfilling the core objectives outlined in Section \ref{sec:problem_def}. By inferring the desired target from the provided finger position and direction, the robot can autonomously initiate actions to reach it.

\subsection{Edge Cases}

We next evaluate the performance of PointingNet in specific but interesting edge cases where accurate predictions may be difficult. Note that previous experiments included some of these edge cases in the test data. However, we provide a designated analysis in this section to better emphasize the abilities of PointingNet. Designated test sets were collected for six edge cases including gloved hands, out-of-frame participants, partly occluded participants, multi-users in the image, pointing while seated and dual-arm lifting. 

\begin{figure}
    \centering
     \includegraphics[width=\linewidth]{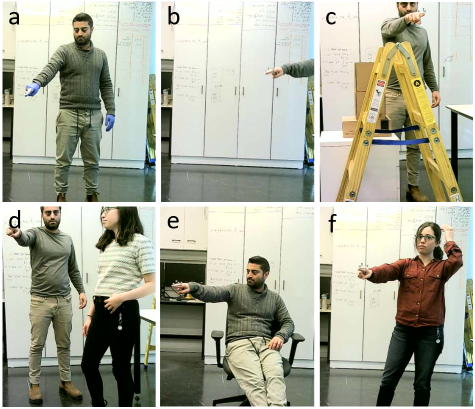} 
    \caption{Examples of the six evaluated pointing edge cases: (a) user wearing gloves, (b) user is out-of-frame and only the pointing arm is visible, (c) user is partly occluded by objects in the scene, (d) multiple people in the scene where the pointing user is in the background, (e) user is sitting down and, (f) both arms of the user are lifted.}
    \label{fig:edge}
\end{figure}

In the gloved hands case, participants pointed while wearing blue or black surgical gloves (Figure \ref{fig:edge}a). This also evaluates model robustness to skin color. In the out-of-frame case, participants were positioned outside the camera's frame with only their pointing arm visible (Figure \ref{fig:edge}b). This demonstrates the independence of the model from other body parts unlike in other methods. Similarly, during occlusion, part of the participant's body is not visible. Hence, the participants stood behind large objects (e.g. a chair and a ladder) with their bodies partially occluded while their pointing arms were clearly visible (Figure \ref{fig:edge}c). In multi-user cases, several participants are seen in the image while only one is pointing (Figure \ref{fig:edge}d). We do not consider cases where more than one participant is pointing. In the fifth case, participants pointed while sitting down (Figure \ref{fig:edge}e). Finally, the dual-arm case involves participants lifting both arms (Figure \ref{fig:edge}f) while one is pointing and the other is performing other tasks (e.g. head scratching or waving). All test sets were collected and labeled as described previously. 
\begin{table}[h]
\centering
\caption{Recognition success rate and estimation accuracy for several edge cases}
\label{tb:edge}
\begin{tabular}{lcccc}\hline
            & Success  & Position  & Pitch  & Yaw  \\
            & (\%)  & (mm) & ($^\circ$) & ($^\circ$) \\\hline
Gloves              & 94.9 & 81.3$\pm$44.0 & 10.89$\pm$6.9 & 11.02$\pm$3.2 \\
Out-of-frame        & 95.5 & 75.5$\pm$35.8 & 10.22$\pm$3.7 & 12.44$\pm$6.1 \\
Occlusions          & 89.5 & 70.9$\pm$29.6 & 11.75$\pm$5.9 & 12.86$\pm$3.7 \\
Multi-users         & 90.5 & 72.3$\pm$34.1 & 6.57$\pm$2.98 & 7.43$\pm$4.54 \\
Sitting-down        & 97.9 & 68.8$\pm$35.6 & 3.87$\pm$1.88 & 5.08$\pm$2.12 \\
Dual-arm            & 87.8 & 70.1$\pm$38.8 & 7.45$\pm$2.61 & 8.03$\pm$2.94 \\
\hline
\end{tabular}
\end{table}

Table \ref{tb:edge} summarizes the classification success rate and pointing estimation for all six edge cases. The success rate remains high with a slight decrease for occlusions, multi-users and dual-arm. Occlusions and out-of-frame are relatively similar scenarios. However, the arm in out-of-frame scenarios is in the foreground and clearly visible. In occlusions, on the other hand, the arm is in the background while foreground objects have the focus. This is likely why occlusion images receive a lower success rate. 

When comparing pointing estimation with the general error, positional errors negligibly increase by 11~mm on average. However, the yaw and pitch errors increase by an average of 8.08$^\circ$ and 7.84$^\circ$, respectively. Estimating pointing direction in these cases is somewhat harder and the mean errors are larger compared to the general error. Nevertheless, these errors remain low and feasible for accurate pointing. One may improve accuracy in these cases by adding corresponding image samples to the training data.

%% file: Experiments.tex
Following the evaluation of PointingNet, the effectiveness and practicality of the model are experimented over robotic platforms in pointing-based human guidance. The user's ability to guide the robot towards a desired goal is an essential aspect of HRI, as it involves direct communication and coordination between the human and the robot. The following experiments provide an assessment of the model's feasibility for real-world applications over two different robotic platforms. The experimental setups are first described followed by results. 
\begin{figure}
    \centering
    \includegraphics[width=\linewidth]{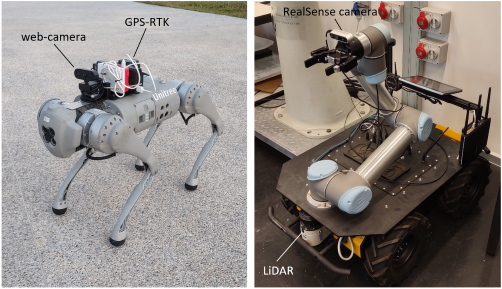} \\
    \begin{tabular}{cc}
         (a) ~~~~~~~~~~~~~~~~~~~~~~~~~~~~ & ~~~~~~~~ (b)  \\
    \end{tabular} 
    \caption{Two robotic platforms used in the experiments: (a) Unitree Go1 quadruped robot equipped with a simple RGB web camera for pointing recognition and a GPS-RTK for localization. (b) Clearpath Husky 4-wheel rover equipped with a RealSense camera (only RGB data is used) for pointing recognition and a 2D LiDAR for localization.}
    \label{fig:robots}
\end{figure}

\subsection{Setup}

\subsubsection{Software} The PointingNet model was wrapped into a Robot Operating System (ROS) node. Hence, the framework described in Figure \ref{fig:robot_scheme} is implemented in ROS and predicts pointing in real-time. The recognition model, with internal PointingNet-S segmentation, loops until pointing is identified. Once identified, the estimation model PointingNet-E is triggered to predict feature vector $\ve{v}$ that corresponds to the captured image. In practice, ROS loops for several more images after the trigger in order for the pointing arm to complete its motion and become static. In our experiments, we assume a flat horizontal floor. Hence, the pointed target is approximated according to \eqref{eq:distance}-\eqref{eq:target_g}. The target is then transferred to a planning and control node which is implemented individually for each of the two robotic platforms described next. In both platforms, however, real-time calculations are performed over a central computer where WiFi communication enables the bi-directional transfer of sensed information and motion commands. 

\subsubsection{Quadruped robot} The first platform is the Unitree Go1 quadruped robot seen in Figure \ref{fig:robots}a. A web camera was mounted on top of the robot to observe pointing users. Furthermore, a GPS-RTK board and an Inertial Measurement Unit (IMU) are used for global localization and heading calculations, respectively. The robot is controlled using the manufacturer's high-level ROS API. Hence, the planning and control node commands the robot to move directly toward the target location and stop once reached. The experimental trials were conducted in an open environment without any obstacles. 

\subsubsection{Husky Rover} The second platform is the Husky 4-wheel rover by Clearpath. The rover is equipped with a Sick 2D LiDAR for localization. Hence, prior to the experiments, the robot mapped the environment to localize itself so that it is able to pinpoint any pointed target within the generated map. To demonstrate the ability of PointingNet to acquire images from various cameras, a RealSense RGB-Depth camera was mounted on the robot as seen in Figure \ref{fig:robots}b. Nevertheless, only RGB data was obtained from the camera. When experimenting with the rover, the pitch angle was ignored. Hence, the robot is instructed to move along a line formed by the projection of the pointed vector on the floor starting from the position of the finger. The ROS node for planning and control will direct the robot to reach the start of the projected line and move along it until encountering an obstacle. In the experimental trials, a small bench was set as the desired target such that the robot could identify it as an obstacle and stop in front of it.

\begin{table}[]
\centering
\caption{Experimental results for robot reaching to pointed targets }
\label{tb:Husky_Go1}
\begin{tabular}{lccc}\toprule
    \multirow{2}{*}{System} & Success & Target reach  & Distance  \\
            & rate (\%) & error (m) & to target (m)  \\\midrule
    Quadruped & 100 & 0.36$\pm$0.28 & 3.8$\pm$0.95 \\
    Rover & 86.7 & 0.32$\pm$0.2 &  7.2$\pm$2.5 \\
\bottomrule
\end{tabular}
\end{table}
\begin{figure}
    \centering
     \includegraphics[width=\linewidth]{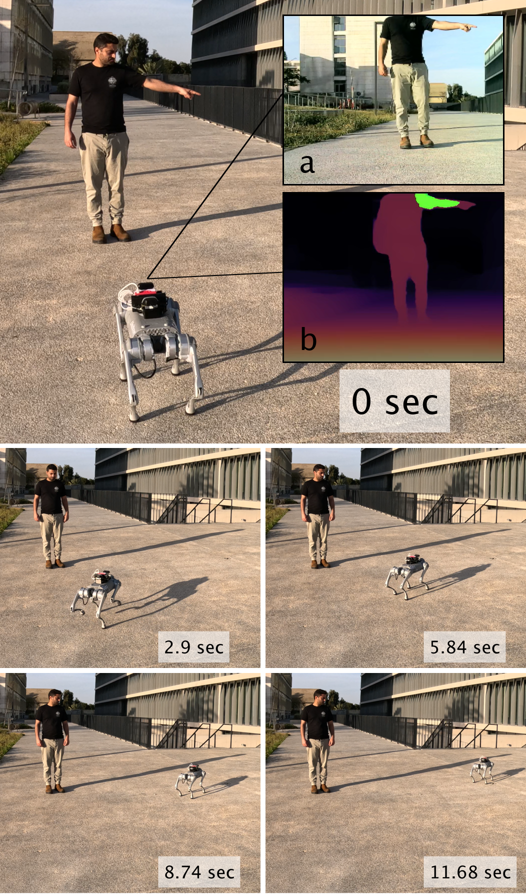} 
    \caption{Snapshots of a pointing experiment with the quadruped robot. The top image shows the moment of pointing along with (a) the image observed by the robot and (b) the estimation of the pointing direction. The robot reached the target with an accuracy of 0.07~m.}
    \label{fig:go1_exp_33}
\end{figure}
\begin{figure}
    \centering
     \includegraphics[width=\linewidth]{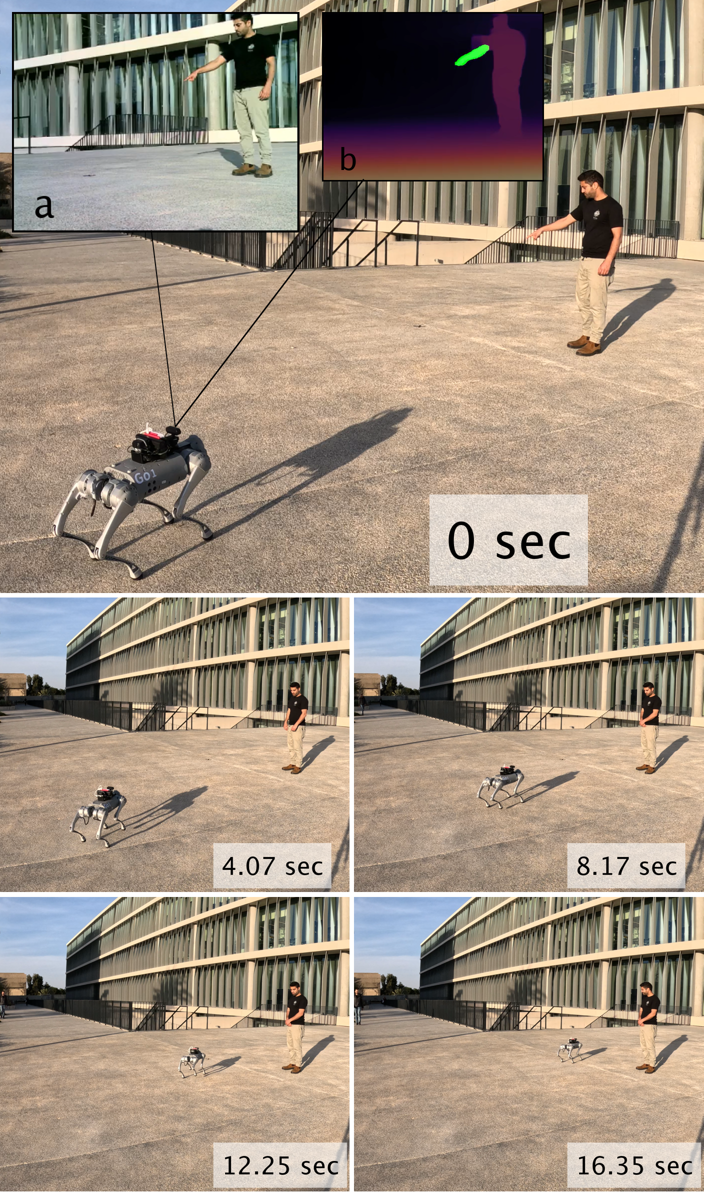} 
    \caption{Snapshots of a pointing experiment with the quadruped robot. The top image shows the moment of pointing along with (a) the image observed by the robot and (b) the estimation of the pointing direction. The robot reached the target with an accuracy of  0.13~m.}
    \label{fig:go1_exp_34}
\end{figure}

\subsection{Results}

A set of experimental trials was conducted for each of the robot platforms. Each set included 15 pointing trials to arbitrary targets with respect to the user and robot. In each trial, the user and robot were randomly relocated such that the user stood in front of the robot at a random relative distance. For the quadruped robot, a random target was marked on the floor. Similarly, a small bench was moved to random positions for the rover to reach. For both robots and in each trial, the user was instructed to point toward the target. 

\begin{figure*}
    \centering
     \includegraphics[width=\linewidth]{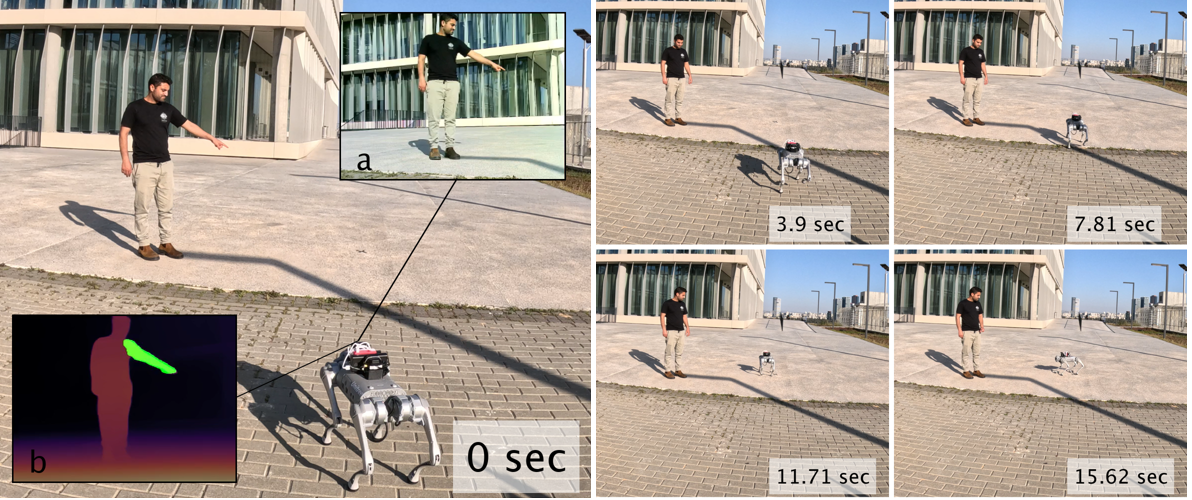} 
    \caption{Snapshots of a pointing experiment with the quadruped robot. The top image shows the moment of pointing along with (a) the image observed by the robot and (b) the estimation of the pointing direction. The robot reached the target with an accuracy of 0.09~m.}
    \label{fig:go1_exp_18}
\end{figure*}

\begin{figure*}
    \centering
     \includegraphics[width=\linewidth]{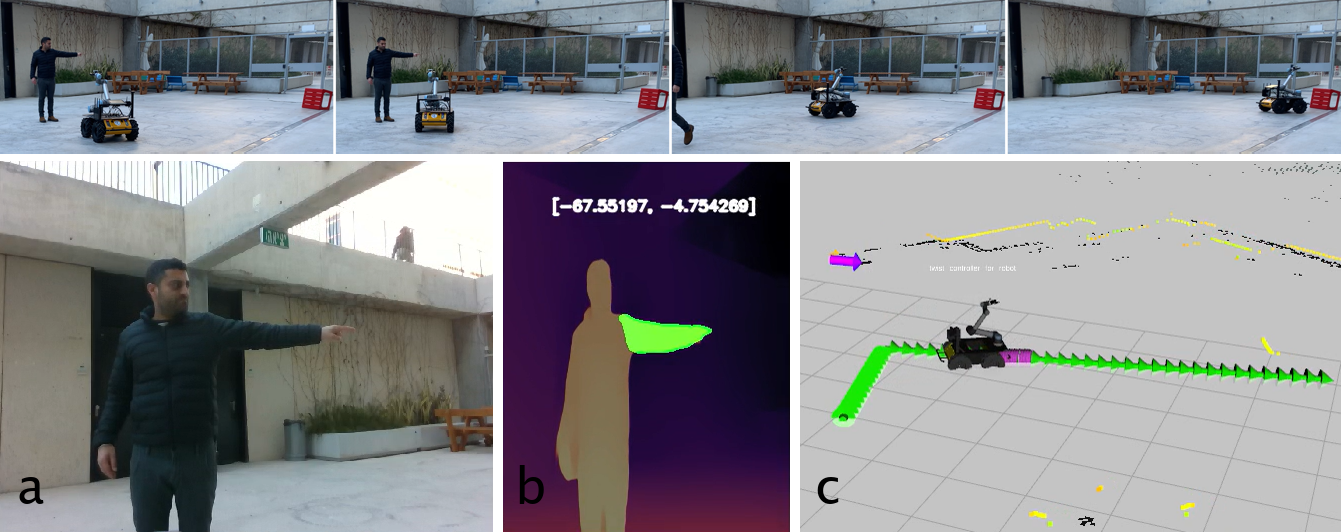} 
    \caption{Pointing experiment with a 4-wheel rover: (top row, left to right) the rover identifies pointing, plans motion and moves towards the target (red bench); (bottom row) (a) image observed by the rover, (b) estimation of pointing direction and (c) planning and controlling the motion in ROS-Gazebo. The purple arrow denotes the position and pointing direction estimation of the index finger with respect to the rover's map and its initial pose. The rover reached the target with an accuracy of 0.3~m.}
    \label{fig:huskey_exp_7}
\end{figure*}

Table \ref{tb:Husky_Go1} summarizes the success rate and mean errors over the 15 pointing trials for each robot. Errors were measured after the robot had stopped at the target and taken from the center of the marked target or bench to the center of the robot. The table also shows the mean distance of the robot from the desired target at the time of pointing. To test various scenarios, the distances for the rover were larger than for the quadruped robot. Figures \ref{fig:go1_exp_33}-\ref{fig:go1_exp_18} and Figure \ref{fig:huskey_exp_7} show several pointing trials of the quadruped robot and rover, respectively. While the quadruped robot reached the vicinity of the target on all trials, the rover missed the bench two times. It is important to note that the reported deviations from the target are also a result of localization errors. The GPS-RTK of the quadruped robot suffered from localization drift and, therefore, errors are larger than for the rover over shorter travel distances. Similarly, some directive routes for the rover had lower mapping features for localization and, consequently, the rover missed two targets. Nevertheless, the results show that the pointing estimation provides sufficient accuracy for the robot to reach a user's desired target through natural pointing and RGB images.  

%% file: Conclusions.tex
This work presented a comprehensive framework enabling a robot to interpret a user’s pointing gesture using an RGB camera and navigate toward the indicated target. Central to the framework is the novel PointingNet model, designed for recognizing pointing gestures and estimating their direction relative to the robot-mounted camera. The PointingNet-S model operates in the background to segment any observed lifted arm, providing focused input for the recognition and estimation models. The combination of arm segmentation by PointingNet-S and MiDaS depth approximation has been demonstrated to be essential for achieving accurate estimations of pointing position and direction. Extensive testing on a large and diverse dataset validated the high accuracy of PointingNet, significantly outperforming state-of-the-art skeleton-based models. The framework also successfully handled challenging edge cases, such as occlusions and multiple participants in the image.

Experimental evaluations on two robotic platforms demonstrated the system's capability to navigate toward human-indicated targets with sufficient accuracy. While the proposed approach generalizes well across different users, environments, and edge cases, its current limitation lies in the effective range of up to five meters. Future work should explore the use of super-resolution techniques to enhance the image quality of pointing gestures observed from greater distances \cite{bamani2024}. Additionally, specific edge cases that yielded higher errors may benefit from targeted data collection to refine the model and improve robustness.

For long-distance pointing, even minor errors in estimation can result in significant deviations from the intended target, leading to ambiguity. To address this, verbal instructions can be integrated into the system to provide additional context about the desired task, enhancing the robot's ability to reach specific objects or locations. For instance, a user could combine pointing with verbal commands such as “green door” or “left-hand side window” to ensure precision. Incorporating such multimodal communication is a promising avenue for future research. Lastly, while pupil gaze is challenging to capture from a distance, future efforts may consider integrating face gaze estimation to provide supplementary directional cues. As discussed in Section \ref{sec:preliminary}, capturing the gaze effectively will require agile processing due to the brief nature of user glances at the target. These enhancements could further extend the applicability and robustness of the proposed framework in real-world scenarios.